\documentclass[pdflatex,sn-mathphys-num]{sn-jnl}

\usepackage{graphicx}%
\usepackage{multirow}%
\usepackage{amsmath,amssymb,amsfonts}%
\usepackage{amsthm}%
\usepackage{mathrsfs}%
\usepackage[title]{appendix}%
\usepackage{xcolor}%
\usepackage{textcomp}%
\usepackage{manyfoot}%
\usepackage{booktabs}%
\usepackage{algorithm}%
\usepackage{algorithmicx}%
\usepackage{algpseudocode}%
\usepackage{listings}%
\usepackage{tikz} 
\usepackage{geometry} 

\theoremstyle{thmstyleone}%
\theoremstyle{thmstyletwo}%

\theoremstyle{thmstylethree}%

\begin{document}

\title[Article Title]{Model Graph Inductive Learning for Knowledge Graph Completion
}


\author[1]{\fnm{Mohommad Esmaeil} \sur{Khani}}\email{mekhani@stu.yazd.ac.ir}
\author[1]{\fnm{Mahdieh} \sur{Hasheminejad}}\email{hasheminezhad@yazd.ac.ir}

\author[2]{\fnm{Ali} \sur{Taherkhani}}\email{
	ali.taherkhani@iasbs.ac.ir}

\author*[3]{\fnm{Hossein} \sur{Hajiabolhassan}}\email{hossein.hajiabolhassan@medunigraz.at}

\affil[1]{\orgdiv{Faculty of Mathematical Science  Department of Computer Science
	}, \orgname{Yazd Unversity}, \orgaddress{,  \postcode{100190}, \country{Iran}}}

\affil[2]{\orgdiv{Department of Mathematics}, \orgname{Institute for Advanced Studies in Basic Sciences (IASBS)}, \orgaddress{  \postcode{45137-66731},  \country{Iran}}}

\affil*[3]{\orgdiv{Diagnostic and Research Institute of Human Genetics}, 
	\orgname{Medizinische Universität Graz}, 
	\orgaddress{\city{Graz}, \postcode{8010}, \country{Austria}}}

%
%

\abstract{
Link prediction in knowledge graphs fundamentally depends on the quality of learned embeddings for entities and relations. However, most existing methods derive these embeddings by aggregating only the local neighborhood of each entity, neglecting the global  structure of the knowledge graph. This limited view prevents models from capturing higher-level structural patterns that are essential for accurate and generalizable link prediction.

To address these limitations, we introduce Model Graph Inductive Learning  (\textbf{MGIL}), a framework that constructs a model graph by clustering entities based on the similarity of their incoming and outgoing relational structures or their entity types. A GNN is then applied to this model graph to produce embeddings that capture the global view of the knowledge graph. These embeddings subsequently serve as high-quality initial features 
for the original knowledge graph, replacing random initialization and leading to more stable and expressive representations.

Extensive experiments on standard and recently proposed inductive benchmarks demonstrate that MGIL achieves state-of-the-art or highly competitive performance in inductive link prediction, highlighting its effectiveness across diverse graph settings.

}




\maketitle

\section{Introduction}\label{sec1}

A knowledge graph (KG) is a heterogeneous graph in which nodes represent entities and edges represent relations that connect them. Knowledge is typically encoded as triples of the form (head, relation, tail). Knowledge graphs provide structured frameworks for representing real-world facts, enabling knowledge inference and logical reasoning. Consequently, they have been widely adopted across many domains to extract and organize structured knowledge from unstructured or semi-structured data.

Building on this structured representation, numerous tasks have 
been formulated for knowledge graphs. Tasks that operate directly 
on the graph structure, such as link prediction \cite{bordes2013transe} 
and triple classification \cite{wang2014transh}, aim to infer missing 
facts or validate existing ones within the graph. In contrast, 
beyond-KG tasks leverage knowledge graphs as auxiliary sources of 
structured knowledge to enhance downstream applications, including 
question answering \cite{yang2015embedding}, recommender systems 
\cite{wang2019knowledge,cao2019unifying,8047276}, and semantic 
web technologies \cite{boroujeni2022answer}.

Real-world knowledge is inherently incomplete and continually 
evolving; as a primary structured representation of such knowledge, 
knowledge graphs are therefore naturally incomplete as well. 
Addressing this incompleteness is crucial, as completing missing 
information enriches knowledge, bridges scientific gaps, and 
enables more advanced reasoning.

Knowledge graph completion, which is most commonly formulated 
as link prediction, seeks to infer missing facts by learning 
structural regularities from the observed portion of the graph 
while preserving real-world semantics. Early transductive embedding 
models, such as TransE \cite{bordes2013transe}, ComplEx 
\cite{trouillon2016complex}, and RotatE \cite{sun2019rotate}, 
embed entities and relations into continuous vector spaces. 
Despite their effectiveness, these models operate in a transductive 
setting, meaning they can only make predictions for entities 
seen during training.

As knowledge graphs grow rapidly, new entities emerge frequently, rendering purely transductive models ineffective in real-world scenarios. As highlighted by Weninger et al. \cite{shi2018open}, the dynamic nature of knowledge demands models capable of generating plausible embeddings for previously unseen entities, that is, inductive models.

Several inductive approaches have been proposed 
\cite{teru2020inductiverelationpredictionsubgraph,xu2022subgraph}; 
however, most of them rely solely on the local neighborhood of 
each entity and initialize embeddings for unseen entities randomly, 
limiting both the quality and generalizability of the learned 
representations. To overcome these limitations, we propose 
\textbf{Model Graph Inductive Learning (MGIL)}, a novel framework 
that generates entity embeddings by integrating both global 
structural patterns of the knowledge graph and the local context 
of individual entities.

Beyond providing a global view of the knowledge graph, the Model Graph also serves as a principled initialization mechanism for entity representations. By assigning entities to structurally coherent clusters, MGIL provides informative initial embeddings that guide the subsequent GNN-based message passing process. This structured initialization facilitates more stable optimization and allows the GNN to learn more meaningful and expressive embeddings for individual entities.

MGIL consists of two tightly coupled components:
\begin{enumerate}
	\item A global module that operates on the model graph to generate 
	structural representations, which serve as meaningful initializations 
	for entity embeddings in subsequent modules (as described in 
	Section~\ref{sec:embeddingLearningTheModelGraph}).
	
	\item A local module that refines these representations using 
	entity-centered induced subgraphs and a graph neural network 
	(see Section~\ref{sec:injectModleGraphEmbedding}).

\end{enumerate}
The final entity embedding is obtained by combining the global prototype (from the model graph) with the locally refined representation. This design yields expressive and generalizable embeddings for both seen and unseen entities in fully inductive settings.

The constructed model graph offers several desirable properties:
\begin{itemize}
\item \textbf{Graph Coarsening:} The model graph contains fewer nodes than the original KG by grouping entities with similar relational patterns or semantic types into abstract nodes, yielding a more compact representation. In some cases, the reduction is substantial.
	
	\item \textbf{Similarity Mapping}: Unseen entities are mapped to the most similar model graph nodes, immediately receiving meaningful embeddings.
	
	\item \textbf{ID-independence}: Since embeddings are derived from structural similarity to nodes of model graph rather than 
	entity identifiers, the model generalizes naturally to 
	unseen entities.
\end{itemize}

Extensive experiments on standard inductive knowledge graph benchmarks show that MGIL consistently achieves state-of-the-art or highly competitive, demonstrating its effectiveness in generalizing to unseen entities.

In summary, our main contributions are as follows:
\begin{enumerate}
	\item The introduction of the Model Graph, a high-level structural abstraction derived from surrounding relational patterns or entity types.
	
\item  The extraction of entity-centered induced subgraphs that enable learning on small local graph structures instead of the entire knowledge graph, leading to more efficient and generalizable modeling.
\item A dual-module framework that jointly learns global 
structural representations and local refinements, enabling 
expressive and generalizable inductive knowledge graph 
completion without relying on entity identifiers.
\end{enumerate}
The source code and datasets used in this paper have been made available at https://github.com/hhaji/MGIL.
\section{Related work} 
Knowledge graph completion and link prediction have attracted 
significant research attention, leading to a diverse range of 
proposed methods. These approaches can broadly be categorized 
into four groups: path-based, embedding-based, rule-based, 
and GNN-based methods.

\subsection{Path-based methods}
Path-based methods exploit graph connectivity patterns to measure similarity between nodes or to infer new links. Originally developed for homogeneous graphs, such methods compute similarity based on various path-related metrics, including the weighted count of paths (Katz index \cite{katz1953new}), random-walk probabilities (personalized PageRank \cite{page1999pagerank}), or shortest-path distances \cite{dijkstra1959note}. SimRank \cite{jeh2002simrank} introduces a measure of similarity based on the expected meeting distance of random walks, which was later extended to heterogeneous graphs by PathSim \cite{sun2011pathsim}. In the context of knowledge graphs, the Path Ranking Algorithm (PRA) \cite{lao2010relational} leverages relational paths as symbolic features for link prediction, enabling reasoning based on explicit multi-hop relations.
Despite their effectiveness, path-based methods suffer from several limitations.
They typically rely on handcrafted features and manually defined paths, which
require domain expertise and extensive feature engineering
\cite{lao2010relational,sun2011pathsim,neelakantan2015compositional}.
Moreover, many of these approaches were originally designed for homogeneous graphs and may require careful design to generalize to complex heterogeneous knowledge graphs.

\subsection{Embedding-based models}
Embedding-based methods aim to represent entities and 
relations of a knowledge graph as low-dimensional vectors 
by training on observed triples, mapping them into a 
continuous vector space where the plausibility of a triple 
$(h, r, t)$ can be measured via a scoring function. 
Classical models such as TransE \cite{bordes2013transe}, 
TransH \cite{wang2014transh}, ComplEx \cite{trouillon2016complex}, 
and RotatE \cite{sun2019rotate} fall into this category. 
These approaches are considered shallow embeddings, 
as they directly assign and optimize embeddings for each 
entity and relation without capturing neighborhood structure 
or inter-triple dependencies. Most of these models operate 
in a transductive setting, meaning they cannot naturally 
generalize to unseen entities. Moreover, incorporating new 
entities or relations often requires full retraining, as 
the model has no mechanism to infer embeddings for 
previously unseen elements.

\subsection{Rule-based methods}
Unlike embedding-based approaches, rule-based methods rely on learning statistical and logical rules by enumerating regularities and patterns present in the knowledge graph. These rules can then be applied for inductive reasoning, enabling generalization to unseen entities. Classical methods such as AMIE \cite{galarraga2013amie} and RuleN \cite{meilicke2018fine} explicitly search for logical rules through discrete exploration, while more recent neural approaches such as Neural LP \cite{yang2017differentiable} and DRUM \cite{sadeghian2019drum} learn rules in an end-to-end differentiable manner. Since the learned rules are independent of specific entities, rule-based models are particularly effective in handling link prediction tasks involving unseen entities.

However, learning a large set of logical rules often introduces scalability challenges, as rule enumeration and evaluation can become computationally expensive on large-scale knowledge graphs. Moreover, the expressiveness of rule-based methods is inherently constrained by predefined rule templates, limiting their ability to capture complex and diverse relational patterns.
\subsection{GNN-based methods}
GNN-based methods leverage the topological structure of knowledge graphs to encode entities and relations into expressive embeddings. Early works such as R-GCN \cite{schlichtkrull2018modeling} and CompGCN \cite{vashishth2020composition} extend standard GNNs, originally designed for simple undirected graphs, to multi-relational knowledge graphs by incorporating relation-specific transformations and relation embeddings. Subgraph-based methods, such as GraIL \cite{teru2020inductiverelationpredictionsubgraph}, further advance this idea by extracting local subgraphs around target triples and scoring them using a GNN, enabling inductive reasoning over unseen entities. Several follow-up works have built on this framework, including \cite{mai2021communicative, xu2022subgraph, an2022cycle}, demonstrating improved performance in link prediction and reasoning tasks.

Despite their strong expressive power, GNN-based methods typically suffer from high computational complexity, particularly when applied to large-scale knowledge graphs with many relations. Moreover, most existing GNN-based approaches generate entity embeddings primarily by exploiting local neighborhood structures. While local subgraphs capture important relational signals, they may overlook global structural characteristics of the knowledge graph, which can provide complementary and high-level relational context.

Another limitation of many GNN-based models is that entity embeddings are commonly initialized using random vectors before message passing. Such random initialization provides no structural prior and may lead to unstable optimization or suboptimal representation learning, especially in inductive settings where entities are unseen during training.

In contrast, our proposed \textbf{Model Graph Inductive Learning} (\textsc{MGIL})  framework explicitly integrates both global and local structural information. MGIL constructs a global \emph{model graph} to capture high-level relational patterns across the entire knowledge graph and uses this abstraction to provide a structured and semantically informed initialization for entity representations. These informed initial representations serve as a principled starting point for subsequent GNN-based message passing, facilitating more stable training and more expressive embeddings. MGIL then combines this global structural signal with local entity-centered subgraphs to generate more robust representations for entities and relations.

\section{Proposed Methodology}

\subsection{Motivation and Main Idea}

A fundamental challenge in knowledge graph modeling lies in the limited ability of conventional methods to generalize to unseen entities at inference time. Most classical representation learning approaches for knowledge graphs~\cite{bordes2013transe,trouillon2016complex,sun2019rotate,wang2014transh,yang2015embedding} operate in  a transductive setting, assuming a fixed set of entities and relations observed during training. Consequently, when new entities emerge at test time, these models cannot naturally produce meaningful latent representations for them, which significantly limits their applicability in dynamic and evolving knowledge graphs.

\begin{figure}[t!]
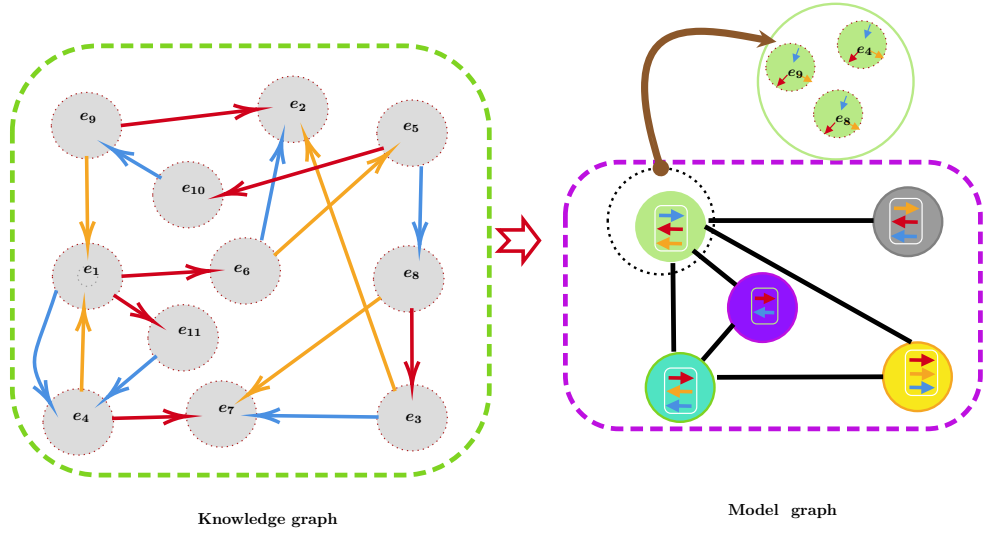

	\centering
\resizebox{\linewidth}{!}{

\tikzset{every picture/.style={line width=0.75pt}} 



}
\caption{Illustration of the model graph construction process from a knowledge graph. 
	In the original knowledge graph (left), entities such as $e_4$, $e_8$, and $e_9$ exhibit similar relational structures,sharing comparable sets of incoming and outgoing relations. 
	In the model graph (right), these structurally similar entities are grouped into a single abstract node (blue). 
	Inside each model-graph node, a bundle of colored arrows represents its relational profile: 
	each color corresponds to a specific relation type, and arrow direction encodes relation orientation. 
	Arrows pointing to the right denote relations whose target entities belong to the cluster, 
	while arrows pointing to the left denote relations whose source entities belong to the cluster.	}

	\label{fig:kg_examples_2}
\end{figure}

A key motivation for our approach is that entities may share high-level structural or semantic similarities even when they are distant or disconnected within the knowledge graph. 
In many cases, two entities can participate in analogous relational patterns and exhibit comparable neighborhood configurations despite having no direct or indirect paths between them. 
As illustrated in Figure~\ref{fig:kg_examples_2}, entities $e_8$ and $e_9$ are located in distant regions of the original knowledge graph, yet they demonstrate highly similar relational interaction patterns. 

\begin{figure}[t!]
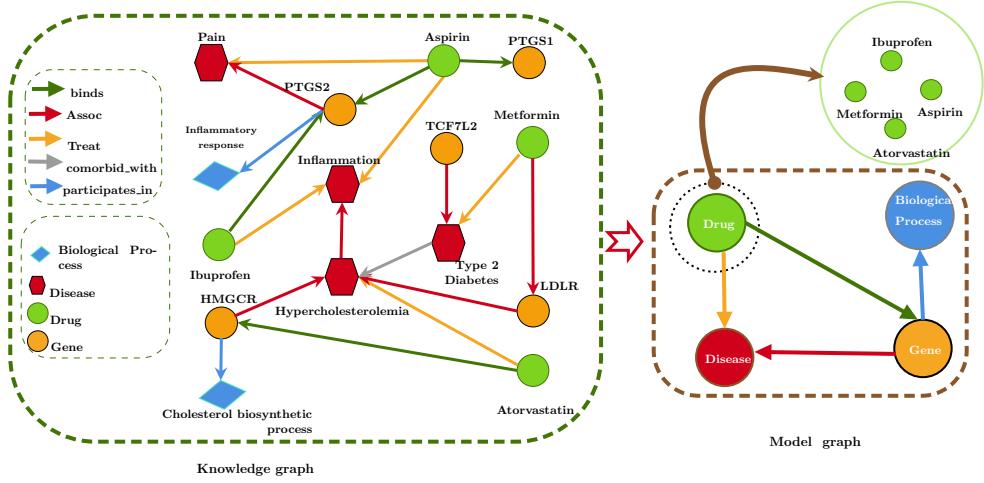

	\centering
	\resizebox{\linewidth}{!}{

\tikzset{every picture/.style={line width=0.75pt}} 



	}
	\caption{The original knowledge graph (left) and the type-based model graph (right), where entities of the same semantic type (e.g., compounds, genes, diseases, biological processes) are aggregated into a single node, with different colors indicating their respective types. 	}
	
	\label{fig:kg_examples_3}
\end{figure}

Moreover, beyond relational pattern similarity, entities may also share intrinsic semantic properties, such as belonging to the same entity type (e.g., drug, protein, or side-effect in biomedical knowledge graphs). Such global structural or semantic commonalities are not necessarily reflected in local connectivity patterns. For example, as illustrated in Figure \ref{fig:kg_examples_3}, entities such as Aspirin and Metformin may not be directly connected in the knowledge graph, yet both belong to the semantic category of drugs. In the corresponding model graph, these entities can therefore be represented by a single shared abstract node representing the drug type.

These observations reveal an important limitation of existing subgraph-based and GNN-based approaches~\cite{teru2020inductiverelationpredictionsubgraph, schlichtkrull2018modeling, vashishth2020composition}, which primarily rely on local message passing over observed graph neighborhoods and therefore may struggle to capture global relational roles or coarse-grained semantic abstractions.

To address this limitation, we construct a model graph that abstracts entities into higher-level nodes according to shared structural characteristics. 
Specifically, entities can be grouped either based on high relational similarity or based on intrinsic semantic types. 
We refer to these two constructions as the \emph{relation-based model graph} and the \emph{entity-based model graph}, respectively. 
The relational construction clusters entities that share analogous incoming and outgoing relation patterns, whereas the type-based construction aggregates entities belonging to the same semantic category into a shared abstract node. 
For example, in a biomedical knowledge graph, all drug entities may be represented by a single abstract node, while all protein entities form another abstract node.

Consider the knowledge graph in Figure~\ref{fig:kg_examples_2}. 
On the left side, the knowledge graph contains entities such as $e_1$, $e_2$, $e_3$, $e_4$, $e_5$, $e_6$, $e_7$, $e_8$, and $e_9$. 
Each entity is associated with a set of incoming and outgoing relations, illustrated by arrows of different colors, where each color denotes a distinct relation type. 

Entities such as $e_4$, $e_8$, and $e_9$ exhibit highly similar relational structures: they share comparable incoming relations (e.g., blue arrows) as well as outgoing relations (e.g., red and orange arrows), together with analogous neighborhood configurations. 
Although these entities may be located in distant regions of the graph, their relational signatures are structurally similar. 
Based on this observation, we construct a \emph{relation-based model graph} in which entities with highly similar relational patterns are grouped into the same abstract node. 
Accordingly, $e_4$, $e_8$, and $e_9$ are mapped into a single cluster (the green node) in the model graph.While this construction relies on relational similarity, an alternative abstraction can be defined based on semantic properties.
Figure~\ref{fig:kg_examples_3} illustrates the construction of an alternative entity-based knowledge graph. Instead of clustering entities based on their relational patterns, entities are grouped according to their intrinsic semantic types. For instance, drugs such as Aspirin, Metformin, and Ibuprofen are represented as a single abstract node, and similarly, all entities of the same type are aggregated into type-specific nodes. This type-based abstraction provides a coarse-grained yet semantically meaningful summarization of the original graph, preserving the overall structure while simplifying visualization and analysis.

This distinction becomes particularly important in large-scale and heterogeneous knowledge graphs such as HetioNet. In such datasets, the graph is often incomplete, and the local neighborhood of many entities may be only partially observed, with several relevant relations missing. Consequently, constructing a model graph purely based on relational patterns may lead to unreliable or fragmented clusters, and in some cases, the number of nodes in the model graph  may approach that of the original graph (see Table~\ref{table:model_graph_stats}).

In these scenarios, an entity-centric abstraction based on intrinsic semantic properties provides a more stable and compact alternative. By grouping entities according to their inherent characteristics rather than relying solely on observed connectivity, the model graph becomes significantly smaller and less sensitive to missing relational information. This leads to a more robust structural representation and can improve generalization performance, particularly in inductive settings where relational evidence is sparse or incomplete.

The resulting model graph is typically smaller than the original knowledge graph, and in some cases substantially smaller, particularly when strong structural or semantic regularities exist among entities. However, this reduction is not guaranteed and depends on the characteristics of the dataset. This compression enables efficient and scalable GNN-based message passing over a more compact abstraction while preserving a global view of relational or semantic patterns.

However, as shown in Table~\ref{table:model_graph_stats}, the degree of compression varies across datasets. In cases where relational patterns are highly diverse or incomplete—such as large-scale heterogeneous graphs (e.g., CoDEx-M or HetioNet)—the reduction in the number of nodes may be less pronounced.

This observation further motivates the use of entity-centric abstractions in certain scenarios. In datasets like HetioNet, where relational neighborhoods may be partially observed or noisy, clustering entities purely based on relational patterns can lead to fragmented representations and relatively large model graphs. In contrast, leveraging intrinsic semantic properties of entities allows for more compact and stable abstractions, significantly reducing the number of model graph nodes and improving representation quality.

\subsection{Model Graph Construction}  \label{sec:graph_construct}
The knowledge graph (KG) is formally defined as 
$\mathcal{G} = (\mathcal{V}, \mathcal{R}, \mathcal{E})$, 
where $\mathcal{V}$ denotes the set of entities, 
$\mathcal{R}$ is the set of relations, and 
$\mathcal{E} \subseteq \mathcal{V} \times \mathcal{R} \times \mathcal{V}$ 
represents the set of edge (i.e relational triples).

The construction of a model graph from a knowledge graph 
involves two main scenarios: relation-based model graphs 
and entity-based model graphs. In a relation-based model 
graph, entities are grouped according to the similarity of 
their relational patterns; each node represents a cluster 
of entities that share similar incoming and outgoing 
relation types. In an entity-based model graph, nodes are 
constructed based on the intrinsic types or features of 
the entities themselves.

\subsubsection{Relation-Based Model Graph} \label{sec:relation-based-model-graph }

We construct the \textit{relation-based model graph} based on the given knowledge graph $\mathcal{G} = (\mathcal{V}, \mathcal{R}, \mathcal{E})$, where $\mathcal{R} = \{r_1, \ldots, r_{|\mathcal{R}|}\}$.

For each entity $v_i \in \mathcal{V}$, we define a relational feature vector $\vec{f}_i \in \{0,1\}^{2|\mathcal{R}|}$ that encodes both outgoing and incoming relations associated with the entity. The first $|\mathcal{R}|$ dimensions correspond to outgoing relations, while the remaining $|\mathcal{R}|$ dimensions correspond to incoming relations. Formally, for each index $j \in \{1, \ldots, 2|\mathcal{R}|\}$:
\begin{equation}
	\vec{f}_i[j] =
	\begin{cases}
		1 & \text{if } j \leq |\mathcal{R}| \text{ and } \exists (v_i, r_j, v_k) \in \mathcal{E}, \\
		1 & \text{if } j > |\mathcal{R}| \text{ and } \exists (v_k, r_{j-|\mathcal{R}|}, v_i) \in \mathcal{E}, \\
		0 & \text{otherwise}.
	\end{cases}
\end{equation}

The duplication of relation dimensions allows us to explicitly distinguish between outgoing and incoming relational roles, which is essential since an entity may participate in both directions of a relation (e.g., parent vs. child).

\begin{figure}[h!]
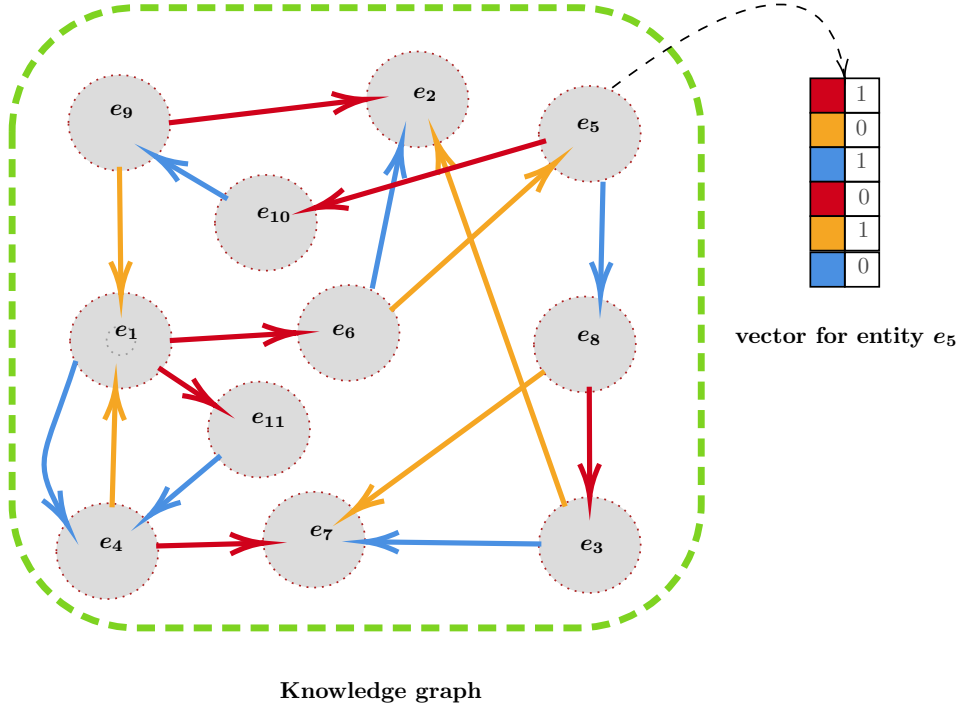

	\centering
\resizebox{\linewidth}{!}{

\tikzset{every picture/.style={line width=0.75pt}} 



}

\caption{Relational feature vector construction for entity $e_5$. Different colors indicate different relation types. The first half of the vector encodes outgoing relations, while the second half represents incoming relations.}

	\label{fig:rel_vec_kg_en}
\end{figure}

Figure~\ref{fig:rel_vec_kg_en} illustrates an example of the relational feature 
vector construction for the entity $e_5$.

After defining the feature vector $\vec{f}_i$ for each entity 
$v_i \in \mathcal{V}$, our goal is to group entities that share 
identical relational feature vectors.

We construct a model graph $M_G = (V_m, E_m)$, where each node 
corresponds to a unique relational feature pattern.

First, we extract the set of unique feature vectors:
\[
F = \left\{ \vec{f}_i \in \{0,1\}^{2|\mathcal{R}|} 
\;\middle|\; v_i \in \mathcal{V} \right\}.
\]

Each unique feature vector $\vec{f}_j \in F$ defines 
a node $U_j \in V_m$ in the model graph:
\[
V_m = \{ U_j \mid \vec{f}_j \in F \}.
\]

Each node $U_j$ represents the set of entities sharing the same 
feature vector:
\[
U_j = \{ v_i \in \mathcal{V} \mid \vec{f}_i = \vec{f}_j \}.
\]

An undirected edge $(U_j, U_k) \in E_m$ is established if there exists at least one triple $(v_p, r, v_q) \in \mathcal{E}$ such that $v_p \in U_j$ and $v_q \in U_k$:
\[
(U_j, U_k) \in E_m \iff \exists (v_p, r, v_q) \in \mathcal{E} \text{ with } v_p \in U_j \text{ and } v_q \in U_k.
\]
Formally, the proposed model graph $M_G$ can be viewed as a 
quotient graph of the original knowledge graph $\mathcal{G}$ 
under an equivalence relation defined over entities, where 
$v_i \sim v_j$ if and only if $\vec{f}_i = \vec{f}_j$. 
Under this formulation, each node in $M_G$ corresponds to an 
equivalence class of entities, and edges capture inter-class 
connectivity induced by the original graph.

The overall construction procedure is summarized in Algorithm~\ref{alg:model-graph}, which describes how relational feature vectors are computed, entities are grouped, and edges are formed based on observed interactions in the knowledge graph.

This construction yields a compact representation of the original knowledge graph, where entities with identical relational roles are aggregated into representative nodes. Such abstraction significantly reduces the graph size while preserving essential structural information, thereby enabling efficient and scalable inductive reasoning.

Table~\ref{table:model_graph_stats} reports the statistics of the constructed model graphs across several benchmark datasets, demonstrating that the proposed construction yields a smaller graph than the original knowledge graph.

\begin{table}[t]
	\centering
	\footnotesize
	\caption{Statistics of various relation model graph versions derived from WN18RR, FB15k-237, and NELL-995. 
		We report the number of entities, relations, triples, nodes, and edges in the constructed model graphs.}
	\label{table:model_graph_stats}
	
	\begin{tabular}{lccccc}
		\toprule
		Dataset & \#Entities & \#Relations & \#Triples & \#Nodes & \#Edges \\
		\midrule
		
		\multicolumn{6}{c}{\textbf{NELL995}} \\
		NELL995\_v1 & 3103 & 14  & 5540  & 408  & 1746 \\
		NELL995\_v2 & 2564 & 88  & 10109 & 750  & 4404 \\
		NELL995\_v3 & 4647 & 142 & 20117 & 1602 & 10446 \\
		NELL995\_v4 & 2092 & 76  & 9289  & 825  & 5274 \\
		
		\midrule
		\multicolumn{6}{c}{\textbf{WN18RR}} \\
		WN18RR\_v1 & 2746  & 9  & 6678  & 128 & 973  \\
		WN18RR\_v2 & 6954  & 10 & 18968 & 210 & 1975 \\
		WN18RR\_v3 & 12078 & 11 & 32150 & 285 & 2726 \\
		WN18RR\_v4 & 3861  & 9  & 9842  & 160 & 1242 \\
		
		\midrule
		\multicolumn{6}{c}{\textbf{FB15k-237}} \\
		FB15k\_v1 & 1594 & 180 & 5226  & 1074 & 4166 \\
		FB15k\_v2 & 2608 & 200 & 12085 & 1815 & 9437 \\
		FB15k\_v3 & 3668 & 215 & 22394 & 2633 & 12633 \\
		FB15k\_v4 & 4707 & 219 & 33916 & 3368 & 25566 \\
		
		\midrule
		\multicolumn{6}{c}{\textbf{Shomer et al.  Datasets \cite{Shomer_2025}}} \\
		CoDEx-M      & 8362  & 47 & 76960  & 7467 & 79727 \\
		HetioNet\_E  & 3898  & 14 & 101667 & 2435 & 99997 \\
		WN18RR\_E    & 12142 & 11 & 24584  & 4346 & 19925 \\
		
		\bottomrule
	\end{tabular}
\end{table}

\begin{algorithm}[htbp]
	\caption{Relation-Based Model Graph Construction}
	\label{alg:model-graph}
	\begin{algorithmic}[1]
		
		\State \textbf{Input:} Knowledge graph $\mathcal{G} = (\mathcal{V}, \mathcal{R}, \mathcal{E})$
		\State \textbf{Output:} Model graph $M_G = (V_m, E_m)$
		
		\State Initialize feature vector $\vec{f}_i = \mathbf{0}$ for all $v_i \in \mathcal{V}$
		
		\ForAll{$(v_i, r_j, v_k) \in \mathcal{E}$}
		\State $\vec{f}_i[j] \gets 1$ \Comment{Outgoing relation}
		\State $\vec{f}_k[j + |\mathcal{R}|] \gets 1$ \Comment{Incoming relation}
		\EndFor
		
		\State \textbf{/* Group nodes with identical feature vectors */}
		\State $V_m \gets \emptyset$
		
		\ForAll{$v_i \in \mathcal{V}$}
		\If{there exists $U \in V_m$ with feature $\vec{f}_i$}
		\State Add $v_i$ to $U$
		\Else
		\State Create new node $U = \{ v_i \}$ and add it to $V_m$
		\EndIf
		\EndFor
		
		\State Initialize $E_m \gets \emptyset$
		
		\State \textbf{/* Define edges in the model graph:
			an edge is created between two nodes if there exists at least one relation between any pair of original nodes in their corresponding groups */}
		
		\ForAll{$(v_i, r, v_j) \in \mathcal{E}$}
		\State Let $U_p, U_q \in V_m$ be the nodes containing $v_i$ and $v_j$
		\If{$U_p \neq U_q$}
		\State Add edge $(U_p, U_q)$ to $E_m$
		\EndIf
		\EndFor
		
		\State \Return $M_G = (V_m, E_m)$
		
	\end{algorithmic}
\end{algorithm}

\subsubsection{Entity-Based Model Graph}
\label{sec:entity-based-model-graph}
To formalize the construction of an \textit{entity-based model 
	graph}, we begin with a given knowledge graph
\[
\mathcal{G} = (\mathcal{V}, \mathcal{R}, \mathcal{E}, \tau),
\]
where $\tau(v)$ denotes the \textit{type} of entity $v \in 
\mathcal{V}$. Let the set of all entity types be denoted as $\{t_1, t_2, \ldots, t_k\}$. The entity-based model graph is defined as $M_G = (V_m, E_m)$, where each node $U_j \in V_m$ corresponds to an entity type $t_j$, for $1 \leq j \leq k$. Specifically, each node $U_j$ represents the subset of entities sharing the same type:
\[
U_j = \{ v_i \in \mathcal{V} \mid \tau(v_i) = t_j \}, 
\quad 1 \leq j \leq k.
\]

We define an undirected edge $(U_j, U_k) \in E_m$ if there exists at least one triple $(h, r, t) \in \mathcal{E}$ such that $\tau(h) = t_j$ and $\tau(t) = t_k$, or vice versa.

This construction can also be interpreted as a quotient graph, 
where the equivalence relation is defined based on entity types, 
i.e., two entities are equivalent if they share the same type.


\subsection{Subgraph Generation Procedure}\label{sec:subgraph_generation}

In this work, we construct a collection of $K$ task-specific 
subgraphs for meta-learning. The subgraph generation procedure 
follows the framework introduced by 
\cite{chen2022metaknowledgetransferinductiveknowledge}, with 
adaptations to ensure consistency with our formulation and to 
support inductive generalization.
Given a knowledge graph $\mathcal{G} = (\mathcal{V}, 
\mathcal{R}, \mathcal{E})$, each subgraph is constructed 
independently through a three-stage sampling process.

First, for each subgraph $i \in \{1, \dots, K\}$, an initial 
entity is randomly sampled from $\mathcal{V}$. Starting from 
this entity, a uniform random walk of length $l_{rw}$ is 
performed $n_{rw}$ times to collect a set of entities, denoted 
as $\mathcal{V}_{rw}^{(i)}$. This procedure is repeated 
$t_{rw}$ times with different starting entities, and the 
resulting sets are merged to improve coverage and structural 
diversity.

Second, an induced subgraph $T_i \subseteq \mathcal{E}$ is 
constructed by including all triples $(h, r, t) \in \mathcal{E}$ 
such that $h, t \in \mathcal{V}_{rw}^{(i)}$:
\[
T_i = \left\{ (h, r, t) \in \mathcal{E} \;\middle|\; 
h, t \in \mathcal{V}_{rw}^{(i)} \right\}.
\]

Third, to prevent information leakage and avoid overfitting to 
entity identities, the entities in each subgraph are anonymized 
by re-indexing them using a random permutation over 
$\{1, \dots, |\mathcal{V}_{rw}^{(i)}|\}$.

Finally, the triples in each subgraph $T_i$ are randomly 
partitioned into two disjoint subsets: a support set $S_i$ and 
a query set $Q_i$, such that $S_i \cup Q_i = T_i$ and 
$S_i \cap Q_i = \emptyset$. The support set is used for model 
adaptation, while the query set is used for evaluation.

The resulting $K$ subgraphs are split into meta-training and 
meta-validation sets using a predefined ratio $\alpha \in (0,1)$, 
where $\lfloor \alpha K \rfloor$ subgraphs are used for 
meta-training and the remainder for validation. For final evaluation, we use the fixed official test triples provided by each benchmark dataset without subgraph sampling. To avoid information leakage, test triples are strictly excluded from the subgraph generation and meta-training process, ensuring that no structural or relational information from the test set is observed during training.

This subgraph-based construction exposes the model to diverse 
structural patterns, improving inductive generalization while 
reducing computational cost compared to training on the full 
graph.

\subsection{Learning Embeddings}

After constructing the model graph and sampling induced subgraphs, the proposed framework learns entity embeddings through three tightly coupled components. 
First, global structural representations are learned on the model graph.
Second, these global embeddings are injected into the knowledge graph and refined using local relational information via an R-GCN.
Finally, the resulting embeddings are evaluated through a scoring function for knowledge graph completion.
We describe each component in detail below.

\subsubsection{Embedding Learning on the Model Graph} \label{sec:embeddingLearningTheModelGraph}

To obtain latent representations for the model graph, each node is
initialized with a random embedding vector sampled from a standard
normal distribution. These 
embeddings are subsequently refined through multiple layers of a 
standard Graph Convolutional Network (GCN), which performs message 
passing and aggregation to capture higher-order structural 
dependencies encoded in the model graph.

Given the model graph $M_G = (V_m, E_m)$, each node
$U \in V_m$ is assigned an initial embedding
$\mathbf{h}_U^{(0)} \in \mathbb{R}^d$. The embeddings are updated
layer by layer as follows:

\begin{equation}
	 \mathbf{h}_U^{(l+1)} = \textbf{ReLU} \left( \sum_{U' \in \mathcal{N}(U) \cup \{U\}} \frac{1}{\sqrt{\tilde{d}_{U'} \tilde{d}_U}} \cdot \mathbf{W}^{(l)} \mathbf{h}_{U'}^{(l)}  \right), \quad U \in V_m, \end{equation}
where $\tilde{d}_U = |\mathcal{N}(U)| + 1$ denotes the degree of node
$U$ after adding self-loops, and $\mathbf{W}^{(l)}$ is a trainable
weight matrix.

After $L$ propagation layers, the embeddings $\mathbf{h}_U^{(L)}$
serve as the final latent representations of nodes in the model graph,
capturing structural interactions among equivalence classes and
encoding high-level structural patterns.

\subsubsection{Injecting Model Graph Embeddings into the Knowledge Graph and Reasoning with R-GCN} \label{sec:injectModleGraphEmbedding}
After obtaining the latent embeddings $\mathbf{h}_U^{(L)}$ for nodes 
in the model graph, these embeddings are used to initialize the 
corresponding entity representations in the knowledge graph. Since 
each node in the model graph may represent multiple entities, the 
embedding of a model node is assigned to all entities mapped to it:

\begin{equation}
	\mathbf{z}_v^{(0)} = \mathbf{h}_{\psi(v)}^{(L)}, \quad \forall v \in \mathcal{V}.
\end{equation}
where $\psi : \mathcal{V} \rightarrow V_m$ is a deterministic mapping 
that assigns each entity to a model graph node according to the 
feature signature used to construct the model graph. Specifically, 
for the relation-based model graph, $\psi(v)$ is determined by the 
relational feature vector of $v$ as described in 
Section~\ref{sec:relation-based-model-graph }; for the entity-based model graph, 
$\psi(v)$ is determined by the entity type or intrinsic feature of $v$ 
as described in Section~\ref{sec:entity-based-model-graph}.

Each node $U \in V_m$ represents a group of entities sharing the same 
feature signature, and the same initial embedding is assigned to all 
entities in the corresponding group.

To refine these initial embeddings, we employ a Relational Graph 
Convolutional Network (R-GCN). At layer $l+1$, the embedding of 
entity $v$ is updated as:

\begin{equation}
	\mathbf{z}_v^{(l+1)} = \textbf{ReLU} \left( 
	\sum_{r \in \mathcal{R}} \sum_{v' \in \mathcal{N}_r(v)} 
	\frac{1}{|\mathcal{N}_r(v)|} \mathbf{W}_r^{(l)} \mathbf{z}_{v'}^{(l)} 
	+ \mathbf{W}_0^{(l)} \mathbf{z}_v^{(l)} 
	\right)
\end{equation}
where $\mathcal{N}_r(v)$ denotes the set of neighbors of $v$ under 
relation $r$, $\mathbf{W}_r^{(l)}$ and $\mathbf{W}_0^{(l)}$ are
trainable relation-specific and self-loop weight matrices, respectively.

Relational reasoning is performed via message passing over sampled 
subgraphs of the knowledge graph. For each subgraph, the initial 
features are injected from the model graph, and refined through 
multiple R-GCN layers. The final embeddings $\mathbf{z}_v^{(L)}$ 
encode latent semantic representations of entities by combining 
high-level structural information from the model graph with local 
relational dependencies captured by the R-GCN.

\subsubsection{Scoring Knowledge Graph Triples}

Once the latent embeddings of entities are obtained, each query triple $(h, r, t)$ is assigned a score using a knowledge graph scoring function. In this work, we adopt TransE as the scoring function, defined as:

\begin{equation}
	s(h, r, t) = -\| \mathbf{z}_h + \mathbf{r}_r - \mathbf{z}_t \|_2
\end{equation}
where $\mathbf{z}_h, \mathbf{z}_t \in \mathbb{R}^d$ denote the learned embeddings of the head and tail entities, and $\mathbf{r}_r \in \mathbb{R}^d$ is the trainable embedding associated with relation $r$. The matrix $\mathbf{R} \in \mathbb{R}^{|\mathcal{R}| \times d}$ contains all relation embeddings.

\medskip

In the meta-training phase, we optimize the model parameters $\theta$ and $\phi$, where $\theta$ corresponds to the model graph encoder and $\phi$ corresponds to the R-GCN-based refinement module. We define a unified embedding function $f_{\text{MG-RGCN}}(\cdot)$, which maps an input subgraph $T_i$ to entity representations by first injecting model graph embeddings and then refining them via relational message passing.

The training objective is defined as:

\begin{equation}
	\mathcal{L}_i = \mathcal{L}(Q_i \mid f_{\text{MG-RGCN}}(T_i), \mathbf{R})
\end{equation}

\medskip
We optimize the model using a self-adversarial negative sampling loss~\cite{sun2019rotate}. For each task $i$ and its query triples $Q_i$, the objective function is defined as:

\begin{equation}
	\begin{aligned}
		\mathcal{L}_i = & \sum_{(h,r,t) \in Q_i} -\log \sigma(\gamma + s(h,r,t)) \\
		& - \sum_{(h',r,t') \in \mathcal{N}_i} p(h',r,t') \log \sigma(-\gamma - s(h',r,t'))
	\end{aligned}
\end{equation}
where $\sigma(\cdot)$ denotes the sigmoid function, $\gamma$ is a fixed margin hyperparameter, and $\mathcal{N}_i$ represents the set of negative samples generated by corrupting either the head or tail entity of each triple.

The self-adversarial weights are computed as:

\begin{equation}
	p(h', r, t') =
	\frac{\exp(\beta \cdot s(h', r, t'))}
	{\sum_{(h'',r,t'') \in \mathcal{N}_i} \exp(\beta \cdot s(h'', r, t''))}
\end{equation}
where $\beta$ is a temperature hyperparameter that controls the sharpness of the distribution and emphasizes harder negative samples by assigning them higher importance during training.

\medskip

After meta-training, the function $f_{\text{MG-RGCN}}$ can generate structure-aware entity embeddings for unseen subgraphs, enabling inductive generalization to new entities and graph structures.

\medskip

\begin{figure}[htbp]
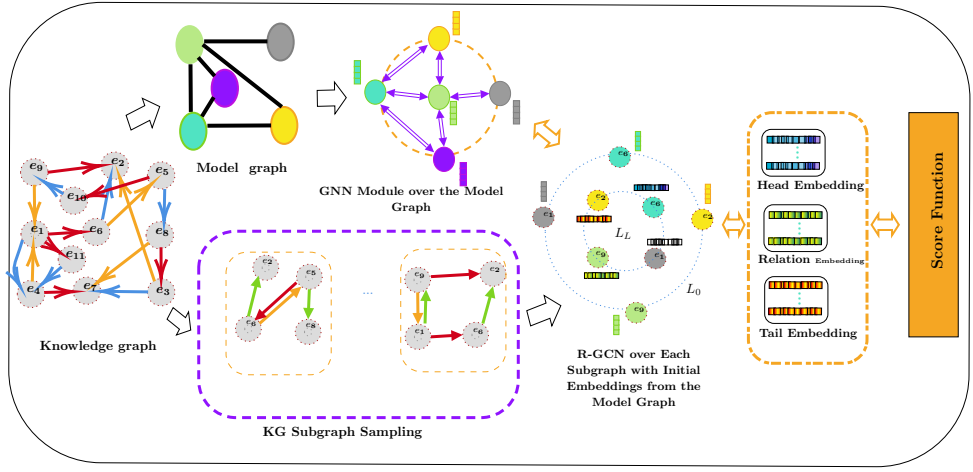


		\scalebox{0.5}{
			\makebox[\textwidth][c]{
				\hspace*{-8cm}


\tikzset {_db7hzz426/.code = {\pgfsetadditionalshadetransform{ \pgftransformshift{\pgfpoint{0 bp } { 0 bp }  }  \pgftransformrotate{0 }  \pgftransformscale{2 }  }}}
\pgfdeclarehorizontalshading{_3aqbtjp97}{150bp}{rgb(0bp)=(1,1,0);
	rgb(37.5bp)=(1,1,0);
	rgb(62.5bp)=(0,0.5,0.5);
	rgb(100bp)=(0,0.5,0.5)}


\tikzset {_owrzn0nor/.code = {\pgfsetadditionalshadetransform{ \pgftransformshift{\pgfpoint{0 bp } { 0 bp }  }  \pgftransformrotate{0 }  \pgftransformscale{2 }  }}}
\pgfdeclarehorizontalshading{_fwta05bat}{150bp}{rgb(0bp)=(1,1,0);
	rgb(37.5bp)=(1,1,0);
	rgb(62.5bp)=(0,0.5,0.5);
	rgb(100bp)=(0,0.5,0.5)}


\tikzset {_gkb0vd16o/.code = {\pgfsetadditionalshadetransform{ \pgftransformshift{\pgfpoint{0 bp } { 0 bp }  }  \pgftransformrotate{0 }  \pgftransformscale{2 }  }}}
\pgfdeclarehorizontalshading{_1brtu24od}{150bp}{rgb(0bp)=(1,1,0);
	rgb(37.5bp)=(1,1,0);
	rgb(62.5bp)=(0,0.5,0.5);
	rgb(100bp)=(0,0.5,0.5)}


\tikzset {_m6a9a7lr3/.code = {\pgfsetadditionalshadetransform{ \pgftransformshift{\pgfpoint{0 bp } { 0 bp }  }  \pgftransformrotate{0 }  \pgftransformscale{2 }  }}}
\pgfdeclarehorizontalshading{_7efpqdm4c}{150bp}{rgb(0bp)=(1,1,0);
	rgb(37.5bp)=(1,1,0);
	rgb(62.5bp)=(0,0.5,0.5);
	rgb(100bp)=(0,0.5,0.5)}


\tikzset {_r57u8slau/.code = {\pgfsetadditionalshadetransform{ \pgftransformshift{\pgfpoint{0 bp } { 0 bp }  }  \pgftransformrotate{0 }  \pgftransformscale{2 }  }}}
\pgfdeclarehorizontalshading{_zvt5l0u7s}{150bp}{rgb(0bp)=(1,1,0);
	rgb(37.5bp)=(1,1,0);
	rgb(62.5bp)=(0,0.5,0.5);
	rgb(100bp)=(0,0.5,0.5)}


\tikzset {_zxgebvmul/.code = {\pgfsetadditionalshadetransform{ \pgftransformshift{\pgfpoint{0 bp } { 0 bp }  }  \pgftransformrotate{0 }  \pgftransformscale{2 }  }}}
\pgfdeclarehorizontalshading{_9o4xfdsd5}{150bp}{rgb(0bp)=(1,1,0);
	rgb(37.5bp)=(1,1,0);
	rgb(62.5bp)=(0,0.5,0.5);
	rgb(100bp)=(0,0.5,0.5)}


\tikzset {_4cmukwnvj/.code = {\pgfsetadditionalshadetransform{ \pgftransformshift{\pgfpoint{0 bp } { 0 bp }  }  \pgftransformrotate{0 }  \pgftransformscale{2 }  }}}
\pgfdeclarehorizontalshading{_lzv3fvp9q}{150bp}{rgb(0bp)=(1,1,0);
	rgb(37.5bp)=(1,1,0);
	rgb(62.5bp)=(0,0.5,0.5);
	rgb(100bp)=(0,0.5,0.5)}


\tikzset {_idcawjknc/.code = {\pgfsetadditionalshadetransform{ \pgftransformshift{\pgfpoint{0 bp } { 0 bp }  }  \pgftransformrotate{0 }  \pgftransformscale{2 }  }}}
\pgfdeclarehorizontalshading{_kyiu61r7c}{150bp}{rgb(0bp)=(1,1,0);
	rgb(37.5bp)=(1,1,0);
	rgb(62.5bp)=(0,0.5,0.5);
	rgb(100bp)=(0,0.5,0.5)}


\tikzset {_x8kzuhi4b/.code = {\pgfsetadditionalshadetransform{ \pgftransformshift{\pgfpoint{0 bp } { 0 bp }  }  \pgftransformrotate{0 }  \pgftransformscale{2 }  }}}
\pgfdeclarehorizontalshading{_6zpujqo88}{150bp}{rgb(0bp)=(1,1,0);
	rgb(37.5bp)=(1,1,0);
	rgb(62.5bp)=(0,0.5,0.5);
	rgb(100bp)=(0,0.5,0.5)}


\tikzset {_bpbsfdebg/.code = {\pgfsetadditionalshadetransform{ \pgftransformshift{\pgfpoint{0 bp } { 0 bp }  }  \pgftransformrotate{0 }  \pgftransformscale{2 }  }}}
\pgfdeclarehorizontalshading{_xi61g6r7d}{150bp}{rgb(0bp)=(1,1,0);
	rgb(37.5bp)=(1,1,0);
	rgb(62.5bp)=(0,0.5,0.5);
	rgb(100bp)=(0,0.5,0.5)}


\tikzset {_ir8cfcj0w/.code = {\pgfsetadditionalshadetransform{ \pgftransformshift{\pgfpoint{0 bp } { 0 bp }  }  \pgftransformrotate{0 }  \pgftransformscale{2 }  }}}
\pgfdeclarehorizontalshading{_rb55naoen}{150bp}{rgb(0bp)=(1,0,0);
	rgb(37.5bp)=(1,0,0);
	rgb(50bp)=(1,1,0);
	rgb(62.5bp)=(1,0,0);
	rgb(100bp)=(1,0,0)}


\tikzset {_d5r5693v9/.code = {\pgfsetadditionalshadetransform{ \pgftransformshift{\pgfpoint{0 bp } { 0 bp }  }  \pgftransformrotate{0 }  \pgftransformscale{2 }  }}}
\pgfdeclarehorizontalshading{_fhok0opqf}{150bp}{rgb(0bp)=(1,0,0);
	rgb(37.5bp)=(1,0,0);
	rgb(50bp)=(1,1,0);
	rgb(62.5bp)=(1,0,0);
	rgb(100bp)=(1,0,0)}


\tikzset {_3l725p409/.code = {\pgfsetadditionalshadetransform{ \pgftransformshift{\pgfpoint{0 bp } { 0 bp }  }  \pgftransformrotate{0 }  \pgftransformscale{2 }  }}}
\pgfdeclarehorizontalshading{_q0bu8k48g}{150bp}{rgb(0bp)=(1,0,0);
	rgb(37.5bp)=(1,0,0);
	rgb(50bp)=(1,1,0);
	rgb(62.5bp)=(1,0,0);
	rgb(100bp)=(1,0,0)}


\tikzset {_kqjjnhguq/.code = {\pgfsetadditionalshadetransform{ \pgftransformshift{\pgfpoint{0 bp } { 0 bp }  }  \pgftransformrotate{0 }  \pgftransformscale{2 }  }}}
\pgfdeclarehorizontalshading{_rrrkkkej3}{150bp}{rgb(0bp)=(1,0,0);
	rgb(37.5bp)=(1,0,0);
	rgb(50bp)=(1,1,0);
	rgb(62.5bp)=(1,0,0);
	rgb(100bp)=(1,0,0)}


\tikzset {_b6c8vn3bv/.code = {\pgfsetadditionalshadetransform{ \pgftransformshift{\pgfpoint{0 bp } { 0 bp }  }  \pgftransformrotate{0 }  \pgftransformscale{2 }  }}}
\pgfdeclarehorizontalshading{_yzmawrhag}{150bp}{rgb(0bp)=(1,0,0);
	rgb(37.5bp)=(1,0,0);
	rgb(50bp)=(1,1,0);
	rgb(62.5bp)=(1,0,0);
	rgb(100bp)=(1,0,0)}


\tikzset {_szfdsfczp/.code = {\pgfsetadditionalshadetransform{ \pgftransformshift{\pgfpoint{0 bp } { 0 bp }  }  \pgftransformrotate{0 }  \pgftransformscale{2 }  }}}
\pgfdeclarehorizontalshading{_ettryvspn}{150bp}{rgb(0bp)=(1,0,0);
	rgb(37.5bp)=(1,0,0);
	rgb(50bp)=(1,1,0);
	rgb(62.5bp)=(1,0,0);
	rgb(100bp)=(1,0,0)}


\tikzset {_vgaabm2a9/.code = {\pgfsetadditionalshadetransform{ \pgftransformshift{\pgfpoint{0 bp } { 0 bp }  }  \pgftransformrotate{0 }  \pgftransformscale{2 }  }}}
\pgfdeclarehorizontalshading{_y3p7m8ic9}{150bp}{rgb(0bp)=(1,0,0);
	rgb(37.5bp)=(1,0,0);
	rgb(50bp)=(1,1,0);
	rgb(62.5bp)=(1,0,0);
	rgb(100bp)=(1,0,0)}


\tikzset {_4l1s9oduf/.code = {\pgfsetadditionalshadetransform{ \pgftransformshift{\pgfpoint{0 bp } { 0 bp }  }  \pgftransformrotate{0 }  \pgftransformscale{2 }  }}}
\pgfdeclarehorizontalshading{_wjxghxruo}{150bp}{rgb(0bp)=(1,0,0);
	rgb(37.5bp)=(1,0,0);
	rgb(50bp)=(1,1,0);
	rgb(62.5bp)=(1,0,0);
	rgb(100bp)=(1,0,0)}


\tikzset {_fcraup4by/.code = {\pgfsetadditionalshadetransform{ \pgftransformshift{\pgfpoint{0 bp } { 0 bp }  }  \pgftransformrotate{0 }  \pgftransformscale{2 }  }}}
\pgfdeclarehorizontalshading{_gwjapbawp}{150bp}{rgb(0bp)=(1,0,0);
	rgb(37.5bp)=(1,0,0);
	rgb(50bp)=(1,1,0);
	rgb(62.5bp)=(1,0,0);
	rgb(100bp)=(1,0,0)}


\tikzset {_9dgces4p8/.code = {\pgfsetadditionalshadetransform{ \pgftransformshift{\pgfpoint{0 bp } { 0 bp }  }  \pgftransformrotate{0 }  \pgftransformscale{2 }  }}}
\pgfdeclarehorizontalshading{_93cme1jmv}{150bp}{rgb(0bp)=(1,0,0);
	rgb(37.5bp)=(1,0,0);
	rgb(50bp)=(1,1,0);
	rgb(62.5bp)=(1,0,0);
	rgb(100bp)=(1,0,0)}


\tikzset {_wdd5v7moh/.code = {\pgfsetadditionalshadetransform{ \pgftransformshift{\pgfpoint{0 bp } { 0 bp }  }  \pgftransformrotate{0 }  \pgftransformscale{2 }  }}}
\pgfdeclarehorizontalshading{_vuyspaxnt}{150bp}{rgb(0bp)=(1,1,1);
	rgb(37.5bp)=(1,1,1);
	rgb(62.5bp)=(0,0,0);
	rgb(100bp)=(0,0,0)}
\tikzset{_x1279d2aa/.code = {\pgfsetadditionalshadetransform{\pgftransformshift{\pgfpoint{0 bp } { 0 bp }  }  \pgftransformrotate{0 }  \pgftransformscale{2 } }}}
\pgfdeclarehorizontalshading{_510tc0enj} {150bp} {color(0bp)=(transparent!0);
	color(37.5bp)=(transparent!0);
	color(62.5bp)=(transparent!10);
	color(100bp)=(transparent!10) } 
\pgfdeclarefading{_6spfnothw}{\tikz \fill[shading=_510tc0enj,_x1279d2aa] (0,0) rectangle (50bp,50bp); } 


\tikzset {_93fe40n26/.code = {\pgfsetadditionalshadetransform{ \pgftransformshift{\pgfpoint{0 bp } { 0 bp }  }  \pgftransformrotate{0 }  \pgftransformscale{2 }  }}}
\pgfdeclarehorizontalshading{_3sdvwf9qe}{150bp}{rgb(0bp)=(1,1,1);
	rgb(37.5bp)=(1,1,1);
	rgb(62.5bp)=(0,0,0);
	rgb(100bp)=(0,0,0)}
\tikzset{_90mamlfdm/.code = {\pgfsetadditionalshadetransform{\pgftransformshift{\pgfpoint{0 bp } { 0 bp }  }  \pgftransformrotate{0 }  \pgftransformscale{2 } }}}
\pgfdeclarehorizontalshading{_b7mvw8c5i} {150bp} {color(0bp)=(transparent!0);
	color(37.5bp)=(transparent!0);
	color(62.5bp)=(transparent!10);
	color(100bp)=(transparent!10) } 
\pgfdeclarefading{_30mgyb90t}{\tikz \fill[shading=_b7mvw8c5i,_90mamlfdm] (0,0) rectangle (50bp,50bp); } 


\tikzset {_pyftdzliy/.code = {\pgfsetadditionalshadetransform{ \pgftransformshift{\pgfpoint{0 bp } { 0 bp }  }  \pgftransformrotate{0 }  \pgftransformscale{2 }  }}}
\pgfdeclarehorizontalshading{_9kc6w8ikq}{150bp}{rgb(0bp)=(1,1,1);
	rgb(37.5bp)=(1,1,1);
	rgb(62.5bp)=(0,0,0);
	rgb(100bp)=(0,0,0)}
\tikzset{_pxkkrt8b5/.code = {\pgfsetadditionalshadetransform{\pgftransformshift{\pgfpoint{0 bp } { 0 bp }  }  \pgftransformrotate{0 }  \pgftransformscale{2 } }}}
\pgfdeclarehorizontalshading{_hgj9yk9ci} {150bp} {color(0bp)=(transparent!0);
	color(37.5bp)=(transparent!0);
	color(62.5bp)=(transparent!10);
	color(100bp)=(transparent!10) } 
\pgfdeclarefading{_xzceg9csr}{\tikz \fill[shading=_hgj9yk9ci,_pxkkrt8b5] (0,0) rectangle (50bp,50bp); } 


\tikzset {_2s52e3lnr/.code = {\pgfsetadditionalshadetransform{ \pgftransformshift{\pgfpoint{0 bp } { 0 bp }  }  \pgftransformrotate{0 }  \pgftransformscale{2 }  }}}
\pgfdeclarehorizontalshading{_595v5qql9}{150bp}{rgb(0bp)=(1,1,1);
	rgb(37.5bp)=(1,1,1);
	rgb(62.5bp)=(0,0,0);
	rgb(100bp)=(0,0,0)}
\tikzset{_6ch8prws6/.code = {\pgfsetadditionalshadetransform{\pgftransformshift{\pgfpoint{0 bp } { 0 bp }  }  \pgftransformrotate{0 }  \pgftransformscale{2 } }}}
\pgfdeclarehorizontalshading{_5yltjhaxb} {150bp} {color(0bp)=(transparent!0);
	color(37.5bp)=(transparent!0);
	color(62.5bp)=(transparent!10);
	color(100bp)=(transparent!10) } 
\pgfdeclarefading{_m2eryf35a}{\tikz \fill[shading=_5yltjhaxb,_6ch8prws6] (0,0) rectangle (50bp,50bp); } 


\tikzset {_exdizml2f/.code = {\pgfsetadditionalshadetransform{ \pgftransformshift{\pgfpoint{0 bp } { 0 bp }  }  \pgftransformrotate{0 }  \pgftransformscale{2 }  }}}
\pgfdeclarehorizontalshading{_ieonipo3m}{150bp}{rgb(0bp)=(1,1,1);
	rgb(37.5bp)=(1,1,1);
	rgb(62.5bp)=(0,0,0);
	rgb(100bp)=(0,0,0)}
\tikzset{_gtq792hx5/.code = {\pgfsetadditionalshadetransform{\pgftransformshift{\pgfpoint{0 bp } { 0 bp }  }  \pgftransformrotate{0 }  \pgftransformscale{2 } }}}
\pgfdeclarehorizontalshading{_ogp88ifva} {150bp} {color(0bp)=(transparent!0);
	color(37.5bp)=(transparent!0);
	color(62.5bp)=(transparent!10);
	color(100bp)=(transparent!10) } 
\pgfdeclarefading{_p5lbx09vu}{\tikz \fill[shading=_ogp88ifva,_gtq792hx5] (0,0) rectangle (50bp,50bp); } 


\tikzset {_c0om4n6vc/.code = {\pgfsetadditionalshadetransform{ \pgftransformshift{\pgfpoint{0 bp } { 0 bp }  }  \pgftransformrotate{0 }  \pgftransformscale{2 }  }}}
\pgfdeclarehorizontalshading{_6tf670bwm}{150bp}{rgb(0bp)=(1,1,1);
	rgb(37.5bp)=(1,1,1);
	rgb(62.5bp)=(0,0,0);
	rgb(100bp)=(0,0,0)}
\tikzset{_y87iatwph/.code = {\pgfsetadditionalshadetransform{\pgftransformshift{\pgfpoint{0 bp } { 0 bp }  }  \pgftransformrotate{0 }  \pgftransformscale{2 } }}}
\pgfdeclarehorizontalshading{_8p7vv4olh} {150bp} {color(0bp)=(transparent!0);
	color(37.5bp)=(transparent!0);
	color(62.5bp)=(transparent!10);
	color(100bp)=(transparent!10) } 
\pgfdeclarefading{_3t950qjw5}{\tikz \fill[shading=_8p7vv4olh,_y87iatwph] (0,0) rectangle (50bp,50bp); } 


\tikzset {_bu3pyftq3/.code = {\pgfsetadditionalshadetransform{ \pgftransformshift{\pgfpoint{0 bp } { 0 bp }  }  \pgftransformrotate{0 }  \pgftransformscale{2 }  }}}
\pgfdeclarehorizontalshading{_hwk4sljy2}{150bp}{rgb(0bp)=(1,1,1);
	rgb(37.5bp)=(1,1,1);
	rgb(62.5bp)=(0,0,0);
	rgb(100bp)=(0,0,0)}
\tikzset{_9hzfoshfm/.code = {\pgfsetadditionalshadetransform{\pgftransformshift{\pgfpoint{0 bp } { 0 bp }  }  \pgftransformrotate{0 }  \pgftransformscale{2 } }}}
\pgfdeclarehorizontalshading{_k3rf2caba} {150bp} {color(0bp)=(transparent!0);
	color(37.5bp)=(transparent!0);
	color(62.5bp)=(transparent!10);
	color(100bp)=(transparent!10) } 
\pgfdeclarefading{_tobx99mr8}{\tikz \fill[shading=_k3rf2caba,_9hzfoshfm] (0,0) rectangle (50bp,50bp); } 


\tikzset {_tzcifuytu/.code = {\pgfsetadditionalshadetransform{ \pgftransformshift{\pgfpoint{0 bp } { 0 bp }  }  \pgftransformrotate{0 }  \pgftransformscale{2 }  }}}
\pgfdeclarehorizontalshading{_oqmagt84s}{150bp}{rgb(0bp)=(1,1,1);
	rgb(37.5bp)=(1,1,1);
	rgb(62.5bp)=(0,0,0);
	rgb(100bp)=(0,0,0)}
\tikzset{_a0exyjm8y/.code = {\pgfsetadditionalshadetransform{\pgftransformshift{\pgfpoint{0 bp } { 0 bp }  }  \pgftransformrotate{0 }  \pgftransformscale{2 } }}}
\pgfdeclarehorizontalshading{_vyr75rofu} {150bp} {color(0bp)=(transparent!0);
	color(37.5bp)=(transparent!0);
	color(62.5bp)=(transparent!10);
	color(100bp)=(transparent!10) } 
\pgfdeclarefading{_igo0vipsg}{\tikz \fill[shading=_vyr75rofu,_a0exyjm8y] (0,0) rectangle (50bp,50bp); } 


\tikzset {_ko0di3j71/.code = {\pgfsetadditionalshadetransform{ \pgftransformshift{\pgfpoint{0 bp } { 0 bp }  }  \pgftransformrotate{0 }  \pgftransformscale{2 }  }}}
\pgfdeclarehorizontalshading{_770r7l9vi}{150bp}{rgb(0bp)=(1,1,1);
	rgb(37.5bp)=(1,1,1);
	rgb(62.5bp)=(0,0,0);
	rgb(100bp)=(0,0,0)}
\tikzset{_gblq7xv62/.code = {\pgfsetadditionalshadetransform{\pgftransformshift{\pgfpoint{0 bp } { 0 bp }  }  \pgftransformrotate{0 }  \pgftransformscale{2 } }}}
\pgfdeclarehorizontalshading{_4re88g7ov} {150bp} {color(0bp)=(transparent!0);
	color(37.5bp)=(transparent!0);
	color(62.5bp)=(transparent!10);
	color(100bp)=(transparent!10) } 
\pgfdeclarefading{_2f9fvoi5w}{\tikz \fill[shading=_4re88g7ov,_gblq7xv62] (0,0) rectangle (50bp,50bp); } 


\tikzset {_a7xw3mv2u/.code = {\pgfsetadditionalshadetransform{ \pgftransformshift{\pgfpoint{0 bp } { 0 bp }  }  \pgftransformrotate{0 }  \pgftransformscale{2 }  }}}
\pgfdeclarehorizontalshading{_xm3cw8cfd}{150bp}{rgb(0bp)=(1,1,1);
	rgb(37.5bp)=(1,1,1);
	rgb(62.5bp)=(0,0,0);
	rgb(100bp)=(0,0,0)}
\tikzset{_ctefapy1h/.code = {\pgfsetadditionalshadetransform{\pgftransformshift{\pgfpoint{0 bp } { 0 bp }  }  \pgftransformrotate{0 }  \pgftransformscale{2 } }}}
\pgfdeclarehorizontalshading{_cmjkaxtkf} {150bp} {color(0bp)=(transparent!0);
	color(37.5bp)=(transparent!0);
	color(62.5bp)=(transparent!10);
	color(100bp)=(transparent!10) } 
\pgfdeclarefading{_f7j4qou4c}{\tikz \fill[shading=_cmjkaxtkf,_ctefapy1h] (0,0) rectangle (50bp,50bp); } 


\tikzset {_eb377o1ys/.code = {\pgfsetadditionalshadetransform{ \pgftransformshift{\pgfpoint{0 bp } { 0 bp }  }  \pgftransformrotate{0 }  \pgftransformscale{2 }  }}}
\pgfdeclarehorizontalshading{_9fji278sc}{150bp}{rgb(0bp)=(1,1,0);
	rgb(37.5bp)=(1,1,0);
	rgb(62.5bp)=(0,0.5,0.5);
	rgb(100bp)=(0,0.5,0.5)}


\tikzset {_k9n1o77yn/.code = {\pgfsetadditionalshadetransform{ \pgftransformshift{\pgfpoint{0 bp } { 0 bp }  }  \pgftransformrotate{0 }  \pgftransformscale{2 }  }}}
\pgfdeclarehorizontalshading{_dneixky3g}{150bp}{rgb(0bp)=(1,1,0);
	rgb(37.5bp)=(1,1,0);
	rgb(62.5bp)=(0,0.5,0.5);
	rgb(100bp)=(0,0.5,0.5)}


\tikzset {_bbgavuvts/.code = {\pgfsetadditionalshadetransform{ \pgftransformshift{\pgfpoint{0 bp } { 0 bp }  }  \pgftransformrotate{0 }  \pgftransformscale{2 }  }}}
\pgfdeclarehorizontalshading{_qhmb2mbrs}{150bp}{rgb(0bp)=(1,1,0);
	rgb(37.5bp)=(1,1,0);
	rgb(62.5bp)=(0,0.5,0.5);
	rgb(100bp)=(0,0.5,0.5)}


\tikzset {_2a50sk499/.code = {\pgfsetadditionalshadetransform{ \pgftransformshift{\pgfpoint{0 bp } { 0 bp }  }  \pgftransformrotate{0 }  \pgftransformscale{2 }  }}}
\pgfdeclarehorizontalshading{_m7fzpncog}{150bp}{rgb(0bp)=(1,1,0);
	rgb(37.5bp)=(1,1,0);
	rgb(62.5bp)=(0,0.5,0.5);
	rgb(100bp)=(0,0.5,0.5)}


\tikzset {_m0gx5t567/.code = {\pgfsetadditionalshadetransform{ \pgftransformshift{\pgfpoint{0 bp } { 0 bp }  }  \pgftransformrotate{0 }  \pgftransformscale{2 }  }}}
\pgfdeclarehorizontalshading{_n7k29t2sl}{150bp}{rgb(0bp)=(1,1,0);
	rgb(37.5bp)=(1,1,0);
	rgb(62.5bp)=(0,0.5,0.5);
	rgb(100bp)=(0,0.5,0.5)}


\tikzset {_ts7sckcsj/.code = {\pgfsetadditionalshadetransform{ \pgftransformshift{\pgfpoint{0 bp } { 0 bp }  }  \pgftransformrotate{0 }  \pgftransformscale{2 }  }}}
\pgfdeclarehorizontalshading{_rv2sy3evn}{150bp}{rgb(0bp)=(1,1,0);
	rgb(37.5bp)=(1,1,0);
	rgb(62.5bp)=(0,0.5,0.5);
	rgb(100bp)=(0,0.5,0.5)}


\tikzset {_fffosg8xs/.code = {\pgfsetadditionalshadetransform{ \pgftransformshift{\pgfpoint{0 bp } { 0 bp }  }  \pgftransformrotate{0 }  \pgftransformscale{2 }  }}}
\pgfdeclarehorizontalshading{_oty29lm7i}{150bp}{rgb(0bp)=(1,1,0);
	rgb(37.5bp)=(1,1,0);
	rgb(62.5bp)=(0,0.5,0.5);
	rgb(100bp)=(0,0.5,0.5)}


\tikzset {_1dciur1f7/.code = {\pgfsetadditionalshadetransform{ \pgftransformshift{\pgfpoint{0 bp } { 0 bp }  }  \pgftransformrotate{0 }  \pgftransformscale{2 }  }}}
\pgfdeclarehorizontalshading{_ke94fob22}{150bp}{rgb(0bp)=(1,1,0);
	rgb(37.5bp)=(1,1,0);
	rgb(62.5bp)=(0,0.5,0.5);
	rgb(100bp)=(0,0.5,0.5)}


\tikzset {_6ijlyymw2/.code = {\pgfsetadditionalshadetransform{ \pgftransformshift{\pgfpoint{0 bp } { 0 bp }  }  \pgftransformrotate{0 }  \pgftransformscale{2 }  }}}
\pgfdeclarehorizontalshading{_m8k3ss9ob}{150bp}{rgb(0bp)=(1,1,0);
	rgb(37.5bp)=(1,1,0);
	rgb(62.5bp)=(0,0.5,0.5);
	rgb(100bp)=(0,0.5,0.5)}


\tikzset {_af05z7osv/.code = {\pgfsetadditionalshadetransform{ \pgftransformshift{\pgfpoint{0 bp } { 0 bp }  }  \pgftransformrotate{0 }  \pgftransformscale{2 }  }}}
\pgfdeclarehorizontalshading{_shjj1gxdz}{150bp}{rgb(0bp)=(1,1,0);
	rgb(37.5bp)=(1,1,0);
	rgb(62.5bp)=(0,0.5,0.5);
	rgb(100bp)=(0,0.5,0.5)}


\tikzset {_mc8ddkz59/.code = {\pgfsetadditionalshadetransform{ \pgftransformshift{\pgfpoint{0 bp } { 0 bp }  }  \pgftransformrotate{0 }  \pgftransformscale{2 }  }}}
\pgfdeclarehorizontalshading{_6b90k205j}{150bp}{rgb(0bp)=(1,1,0);
	rgb(37.5bp)=(1,1,0);
	rgb(62.5bp)=(0,0.5,0.5);
	rgb(100bp)=(0,0.5,0.5)}


\tikzset {_6xtem7pu9/.code = {\pgfsetadditionalshadetransform{ \pgftransformshift{\pgfpoint{0 bp } { 0 bp }  }  \pgftransformrotate{0 }  \pgftransformscale{2 }  }}}
\pgfdeclarehorizontalshading{_ywim7ze27}{150bp}{rgb(0bp)=(1,1,0);
	rgb(37.5bp)=(1,1,0);
	rgb(62.5bp)=(0,0.5,0.5);
	rgb(100bp)=(0,0.5,0.5)}


\tikzset {_hwuz2uikv/.code = {\pgfsetadditionalshadetransform{ \pgftransformshift{\pgfpoint{0 bp } { 0 bp }  }  \pgftransformrotate{0 }  \pgftransformscale{2 }  }}}
\pgfdeclarehorizontalshading{_3imsl934i}{150bp}{rgb(0bp)=(1,1,0);
	rgb(37.5bp)=(1,1,0);
	rgb(62.5bp)=(0,0.5,0.5);
	rgb(100bp)=(0,0.5,0.5)}


\tikzset {_j9f4do996/.code = {\pgfsetadditionalshadetransform{ \pgftransformshift{\pgfpoint{0 bp } { 0 bp }  }  \pgftransformrotate{0 }  \pgftransformscale{2 }  }}}
\pgfdeclarehorizontalshading{_dhskwszn5}{150bp}{rgb(0bp)=(1,1,0);
	rgb(37.5bp)=(1,1,0);
	rgb(62.5bp)=(0,0.5,0.5);
	rgb(100bp)=(0,0.5,0.5)}


\tikzset {_end805g6h/.code = {\pgfsetadditionalshadetransform{ \pgftransformshift{\pgfpoint{0 bp } { 0 bp }  }  \pgftransformrotate{0 }  \pgftransformscale{2 }  }}}
\pgfdeclarehorizontalshading{_nwojyfv9o}{150bp}{rgb(0bp)=(1,1,0);
	rgb(37.5bp)=(1,1,0);
	rgb(62.5bp)=(0,0.5,0.5);
	rgb(100bp)=(0,0.5,0.5)}


\tikzset {_0zpjqrxax/.code = {\pgfsetadditionalshadetransform{ \pgftransformshift{\pgfpoint{0 bp } { 0 bp }  }  \pgftransformrotate{0 }  \pgftransformscale{2 }  }}}
\pgfdeclarehorizontalshading{_egrb7b4sp}{150bp}{rgb(0bp)=(1,1,0);
	rgb(37.5bp)=(1,1,0);
	rgb(62.5bp)=(0,0.5,0.5);
	rgb(100bp)=(0,0.5,0.5)}


\tikzset {_h2nnoymqc/.code = {\pgfsetadditionalshadetransform{ \pgftransformshift{\pgfpoint{0 bp } { 0 bp }  }  \pgftransformrotate{0 }  \pgftransformscale{2 }  }}}
\pgfdeclarehorizontalshading{_ai02wkz16}{150bp}{rgb(0bp)=(1,1,0);
	rgb(37.5bp)=(1,1,0);
	rgb(62.5bp)=(0,0.5,0.5);
	rgb(100bp)=(0,0.5,0.5)}


\tikzset {_sn94f56aa/.code = {\pgfsetadditionalshadetransform{ \pgftransformshift{\pgfpoint{0 bp } { 0 bp }  }  \pgftransformrotate{0 }  \pgftransformscale{2 }  }}}
\pgfdeclarehorizontalshading{_r1nph60pn}{150bp}{rgb(0bp)=(1,1,0);
	rgb(37.5bp)=(1,1,0);
	rgb(62.5bp)=(0,0.5,0.5);
	rgb(100bp)=(0,0.5,0.5)}


\tikzset {_jocq3oyvq/.code = {\pgfsetadditionalshadetransform{ \pgftransformshift{\pgfpoint{0 bp } { 0 bp }  }  \pgftransformrotate{0 }  \pgftransformscale{2 }  }}}
\pgfdeclarehorizontalshading{_kfcrin70j}{150bp}{rgb(0bp)=(1,1,0);
	rgb(37.5bp)=(1,1,0);
	rgb(62.5bp)=(0,0.5,0.5);
	rgb(100bp)=(0,0.5,0.5)}


\tikzset {_ohqw800mk/.code = {\pgfsetadditionalshadetransform{ \pgftransformshift{\pgfpoint{0 bp } { 0 bp }  }  \pgftransformrotate{0 }  \pgftransformscale{2 }  }}}
\pgfdeclarehorizontalshading{_2ysz3mit1}{150bp}{rgb(0bp)=(1,1,0);
	rgb(37.5bp)=(1,1,0);
	rgb(62.5bp)=(0,0.5,0.5);
	rgb(100bp)=(0,0.5,0.5)}


\tikzset {_pbqmdaowh/.code = {\pgfsetadditionalshadetransform{ \pgftransformshift{\pgfpoint{0 bp } { 0 bp }  }  \pgftransformrotate{0 }  \pgftransformscale{2 }  }}}
\pgfdeclarehorizontalshading{_qcihxvt17}{150bp}{rgb(0bp)=(1,0,0);
	rgb(37.5bp)=(1,0,0);
	rgb(50bp)=(1,1,0);
	rgb(62.5bp)=(1,0,0);
	rgb(100bp)=(1,0,0)}


\tikzset {_nnu7brhmd/.code = {\pgfsetadditionalshadetransform{ \pgftransformshift{\pgfpoint{0 bp } { 0 bp }  }  \pgftransformrotate{0 }  \pgftransformscale{2 }  }}}
\pgfdeclarehorizontalshading{_znwij3wma}{150bp}{rgb(0bp)=(1,0,0);
	rgb(37.5bp)=(1,0,0);
	rgb(50bp)=(1,1,0);
	rgb(62.5bp)=(1,0,0);
	rgb(100bp)=(1,0,0)}


\tikzset {_io5q2yoie/.code = {\pgfsetadditionalshadetransform{ \pgftransformshift{\pgfpoint{0 bp } { 0 bp }  }  \pgftransformrotate{0 }  \pgftransformscale{2 }  }}}
\pgfdeclarehorizontalshading{_zw8s042ed}{150bp}{rgb(0bp)=(1,0,0);
	rgb(37.5bp)=(1,0,0);
	rgb(50bp)=(1,1,0);
	rgb(62.5bp)=(1,0,0);
	rgb(100bp)=(1,0,0)}


\tikzset {_0csd6pcsi/.code = {\pgfsetadditionalshadetransform{ \pgftransformshift{\pgfpoint{0 bp } { 0 bp }  }  \pgftransformrotate{0 }  \pgftransformscale{2 }  }}}
\pgfdeclarehorizontalshading{_cieuikpqy}{150bp}{rgb(0bp)=(1,0,0);
	rgb(37.5bp)=(1,0,0);
	rgb(50bp)=(1,1,0);
	rgb(62.5bp)=(1,0,0);
	rgb(100bp)=(1,0,0)}


\tikzset {_s3qqdpjfc/.code = {\pgfsetadditionalshadetransform{ \pgftransformshift{\pgfpoint{0 bp } { 0 bp }  }  \pgftransformrotate{0 }  \pgftransformscale{2 }  }}}
\pgfdeclarehorizontalshading{_rfzkszd4j}{150bp}{rgb(0bp)=(1,0,0);
	rgb(37.5bp)=(1,0,0);
	rgb(50bp)=(1,1,0);
	rgb(62.5bp)=(1,0,0);
	rgb(100bp)=(1,0,0)}


\tikzset {_r13zyr9cm/.code = {\pgfsetadditionalshadetransform{ \pgftransformshift{\pgfpoint{0 bp } { 0 bp }  }  \pgftransformrotate{0 }  \pgftransformscale{2 }  }}}
\pgfdeclarehorizontalshading{_nzgspxpep}{150bp}{rgb(0bp)=(1,0,0);
	rgb(37.5bp)=(1,0,0);
	rgb(50bp)=(1,1,0);
	rgb(62.5bp)=(1,0,0);
	rgb(100bp)=(1,0,0)}


\tikzset {_e2zlkhcqw/.code = {\pgfsetadditionalshadetransform{ \pgftransformshift{\pgfpoint{0 bp } { 0 bp }  }  \pgftransformrotate{0 }  \pgftransformscale{2 }  }}}
\pgfdeclarehorizontalshading{_lh7q05akt}{150bp}{rgb(0bp)=(1,0,0);
	rgb(37.5bp)=(1,0,0);
	rgb(50bp)=(1,1,0);
	rgb(62.5bp)=(1,0,0);
	rgb(100bp)=(1,0,0)}


\tikzset {_1n13tuy47/.code = {\pgfsetadditionalshadetransform{ \pgftransformshift{\pgfpoint{0 bp } { 0 bp }  }  \pgftransformrotate{0 }  \pgftransformscale{2 }  }}}
\pgfdeclarehorizontalshading{_oscnizoge}{150bp}{rgb(0bp)=(1,0,0);
	rgb(37.5bp)=(1,0,0);
	rgb(50bp)=(1,1,0);
	rgb(62.5bp)=(1,0,0);
	rgb(100bp)=(1,0,0)}


\tikzset {_yr1onew3p/.code = {\pgfsetadditionalshadetransform{ \pgftransformshift{\pgfpoint{0 bp } { 0 bp }  }  \pgftransformrotate{0 }  \pgftransformscale{2 }  }}}
\pgfdeclarehorizontalshading{_jutzc3tbi}{150bp}{rgb(0bp)=(1,0,0);
	rgb(37.5bp)=(1,0,0);
	rgb(50bp)=(1,1,0);
	rgb(62.5bp)=(1,0,0);
	rgb(100bp)=(1,0,0)}


\tikzset {_cmhd6jutz/.code = {\pgfsetadditionalshadetransform{ \pgftransformshift{\pgfpoint{0 bp } { 0 bp }  }  \pgftransformrotate{0 }  \pgftransformscale{2 }  }}}
\pgfdeclarehorizontalshading{_1ssfge3er}{150bp}{rgb(0bp)=(1,0,0);
	rgb(37.5bp)=(1,0,0);
	rgb(50bp)=(1,1,0);
	rgb(62.5bp)=(1,0,0);
	rgb(100bp)=(1,0,0)}


\tikzset {_6izbkbal1/.code = {\pgfsetadditionalshadetransform{ \pgftransformshift{\pgfpoint{0 bp } { 0 bp }  }  \pgftransformrotate{0 }  \pgftransformscale{2 }  }}}
\pgfdeclarehorizontalshading{_2287hsdd5}{150bp}{rgb(0bp)=(1,0,0);
	rgb(37.5bp)=(1,0,0);
	rgb(50bp)=(1,1,0);
	rgb(62.5bp)=(1,0,0);
	rgb(100bp)=(1,0,0)}


\tikzset {_kyfgtwa8o/.code = {\pgfsetadditionalshadetransform{ \pgftransformshift{\pgfpoint{0 bp } { 0 bp }  }  \pgftransformrotate{0 }  \pgftransformscale{2 }  }}}
\pgfdeclarehorizontalshading{_rs6zretdl}{150bp}{rgb(0bp)=(1,0,0);
	rgb(37.5bp)=(1,0,0);
	rgb(50bp)=(1,1,0);
	rgb(62.5bp)=(1,0,0);
	rgb(100bp)=(1,0,0)}


\tikzset {_qhy89y22w/.code = {\pgfsetadditionalshadetransform{ \pgftransformshift{\pgfpoint{0 bp } { 0 bp }  }  \pgftransformrotate{0 }  \pgftransformscale{2 }  }}}
\pgfdeclarehorizontalshading{_i2lb2y49w}{150bp}{rgb(0bp)=(1,0,0);
	rgb(37.5bp)=(1,0,0);
	rgb(50bp)=(1,1,0);
	rgb(62.5bp)=(1,0,0);
	rgb(100bp)=(1,0,0)}


\tikzset {_1a0g7c8hq/.code = {\pgfsetadditionalshadetransform{ \pgftransformshift{\pgfpoint{0 bp } { 0 bp }  }  \pgftransformrotate{0 }  \pgftransformscale{2 }  }}}
\pgfdeclarehorizontalshading{_nkvflgk0k}{150bp}{rgb(0bp)=(1,0,0);
	rgb(37.5bp)=(1,0,0);
	rgb(50bp)=(1,1,0);
	rgb(62.5bp)=(1,0,0);
	rgb(100bp)=(1,0,0)}


\tikzset {_nethku3h9/.code = {\pgfsetadditionalshadetransform{ \pgftransformshift{\pgfpoint{0 bp } { 0 bp }  }  \pgftransformrotate{0 }  \pgftransformscale{2 }  }}}
\pgfdeclarehorizontalshading{_wnujnj1zq}{150bp}{rgb(0bp)=(1,0,0);
	rgb(37.5bp)=(1,0,0);
	rgb(50bp)=(1,1,0);
	rgb(62.5bp)=(1,0,0);
	rgb(100bp)=(1,0,0)}


\tikzset {_p2aweqnon/.code = {\pgfsetadditionalshadetransform{ \pgftransformshift{\pgfpoint{0 bp } { 0 bp }  }  \pgftransformrotate{0 }  \pgftransformscale{2 }  }}}
\pgfdeclarehorizontalshading{_922ij3rs0}{150bp}{rgb(0bp)=(1,0,0);
	rgb(37.5bp)=(1,0,0);
	rgb(50bp)=(1,1,0);
	rgb(62.5bp)=(1,0,0);
	rgb(100bp)=(1,0,0)}


\tikzset {_m4dpohc6x/.code = {\pgfsetadditionalshadetransform{ \pgftransformshift{\pgfpoint{0 bp } { 0 bp }  }  \pgftransformrotate{0 }  \pgftransformscale{2 }  }}}
\pgfdeclarehorizontalshading{_0zmq034xb}{150bp}{rgb(0bp)=(1,0,0);
	rgb(37.5bp)=(1,0,0);
	rgb(50bp)=(1,1,0);
	rgb(62.5bp)=(1,0,0);
	rgb(100bp)=(1,0,0)}


\tikzset {_hzzzkirsr/.code = {\pgfsetadditionalshadetransform{ \pgftransformshift{\pgfpoint{0 bp } { 0 bp }  }  \pgftransformrotate{0 }  \pgftransformscale{2 }  }}}
\pgfdeclarehorizontalshading{_ogp3p5p3z}{150bp}{rgb(0bp)=(1,0,0);
	rgb(37.5bp)=(1,0,0);
	rgb(50bp)=(1,1,0);
	rgb(62.5bp)=(1,0,0);
	rgb(100bp)=(1,0,0)}


\tikzset {_9ey5nszc5/.code = {\pgfsetadditionalshadetransform{ \pgftransformshift{\pgfpoint{0 bp } { 0 bp }  }  \pgftransformrotate{0 }  \pgftransformscale{2 }  }}}
\pgfdeclarehorizontalshading{_mvdcbxi6n}{150bp}{rgb(0bp)=(1,0,0);
	rgb(37.5bp)=(1,0,0);
	rgb(50bp)=(1,1,0);
	rgb(62.5bp)=(1,0,0);
	rgb(100bp)=(1,0,0)}


\tikzset {_jxys6knov/.code = {\pgfsetadditionalshadetransform{ \pgftransformshift{\pgfpoint{0 bp } { 0 bp }  }  \pgftransformrotate{0 }  \pgftransformscale{2 }  }}}
\pgfdeclarehorizontalshading{_fswt38s3t}{150bp}{rgb(0bp)=(1,0,0);
	rgb(37.5bp)=(1,0,0);
	rgb(50bp)=(1,1,0);
	rgb(62.5bp)=(1,0,0);
	rgb(100bp)=(1,0,0)}
\tikzset{every picture/.style={line width=0.75pt}} 



				}
			
		}
		
		\caption{Overview of the proposed framework. The input knowledge graph is first transformed into a model graph. The model graph is then processed using a multi-layer Graph Neural Network (GNN) to extract latent embeddings for each node. These embeddings are injected into the corresponding entities in sampled subgraphs of the knowledge graph. Finally, an R-GCN is applied to each subgraph to obtain the final entity embeddings.}
		\label{fig:embedding-pipeline}
		
	\end{figure}

The overall pipeline of the proposed framework is illustrated in Figure~\ref{fig:embedding-pipeline}, where the input knowledge graph is first transformed into a model graph, then encoded using a multi-layer GNN, and finally injected into the knowledge graph for R-GCN-based reasoning.

%
%
%
%

\section{Experimental Setup}

In this section, we evaluate the effectiveness of \textsc{MGIL} on inductive link prediction. 
Models are trained on a source knowledge graph and evaluated on an entity-disjoint target graph containing entities unseen during training.

We evaluate \textsc{MGIL} on two categories of benchmark datasets. 
The first follows the inductive splits introduced by \citet{teru2020inductiverelationpredictionsubgraph} and \citet{mai2021communicative}. 
The second consists of datasets proposed by \citet{Shomer_2025}, which minimize structural overlap between source and target graphs to provide a stricter test of inductive generalization.

During training, we construct task-specific subgraphs from the source KG using random-walk-based sampling (detailed in Section~\ref{sec:subgraph_generation}). 
Each task consists of a sampled subgraph used to learn transferable structural representations.

For each dataset, \textsc{MGIL} is trained on the source KG and evaluated on 
the target KG by predicting missing triples. No additional subgraph sampling 
is performed during evaluation; performance is measured directly on the full 
target graph. This evaluation protocol enables a systematic examination of the 
model’s capacity to generalize structural and relational information to unseen 
entities.
\subsection{Constructing the Model Graph}

We construct two types of model graphs depending on the availability of entity type information in the knowledge graph:

\paragraph{Relation-Based Model Graph.} This construction requires no additional entity type information and can be applied to any knowledge graph. Each node represents a unique relation type from the source knowledge graph, and edges capture co-occurrence patterns between relations. We use this approach for all benchmark datasets. Table~\ref{table:model_graph_stats} summarizes the statistics of the constructed relation-based model graphs, including the number of entities and relations in the original knowledge graph, as well as the corresponding nodes and edges in the model graph.

\paragraph{Entity-Based Model Graph.} This construction requires explicit entity type annotations. Each node represents an entity type, and edges capture relational patterns between entity types. We apply this approach to the Hetionet dataset from \citet{Shomer_2025}, a biomedical knowledge graph where detailed entity type information is available.

%
%
%
%
%
\subsection{Task Sampling Strategy}
We adopt the subgraph sampling strategy detailed in Section~\ref{sec:subgraph_generation}, inspired by \textsc{MorsE}~\citep{chen2022metaknowledgetransferinductiveknowledge}. This episodic construction is applied during both training and evaluation.

\subsection{Implementation Details}
We implement \textsc{MGIL} using PyTorch~\cite{paszke2019pytorch} and DGL~\cite{wang2019dgl}. For the inductive link prediction task, we utilize a combination of a GNN for node embeddings and an R-GCN applied to the task graph. To ensure robustness, all reported results are averaged over 5 independent runs with different  seeds. The low variance across runs confirms the stability of our method. \textbf{Detailed hyperparameters, hardware specifications, and baseline configurations are provided in Appendix \ref{sec:appendix_implementation}.}

\subsection{Datasets}

We evaluate \textsc{MGIL} on two groups of inductive benchmark datasets.

\textbf{Standard inductive benchmarks.}
The first group comprises inductive variants of WN18RR~\cite{dettmers2018conve}, FB15k-237~\cite{toutanova2015observed}, and NELL-995~\cite{xiong2017deeppath}, constructed following the inductive evaluation protocol proposed by \citet{teru2020inductiverelationpredictionsubgraph}. 
Under this protocol, the original knowledge graph is partitioned into two entity-disjoint subgraphs: a \emph{source graph} used for training and a \emph{target graph} used for evaluation. The target graph contains entities unseen during training, requiring models to generalize inductively. 
Each benchmark provides four variants (v1–v4) with increasing graph sparsity and structural complexity.

\textbf{Recently proposed inductive benchmarks.}
The second group consists of three benchmarks introduced by \citet{Shomer_2025}: CoDEx-M, WN18RR (Shomer et al. split), and HetioNet. 
These benchmarks are specifically designed to provide a stricter inductive evaluation setting by ensuring that entities in the test set are disjoint from those observed during training, while also reducing structural overlap between the training and test graphs.

As discussed by \citet{Shomer_2025}, such design choices mitigate information leakage and limit the ability of models to rely on memorization of entity-specific patterns. 
Instead, models are required to generalize based on relational and structural patterns present in the graph.

Therefore, these datasets offer a more realistic and challenging benchmark for inductive link prediction, and are particularly well aligned with our approach, which emphasizes learning transferable structural representations through the model graph.

\subsection{Evaluation Protocol}

We evaluate the performance of \textsc{MGIL} on the inductive link prediction task. In this setting, entities appearing in the test set may not have been observed during training, and their embeddings must be constructed at inference time.

\paragraph{Inductive entity embedding.}
For unseen entities in the test set, embeddings are generated using the relation-based model graph. Specifically, for a given entity, we first extract its incident relations (both incoming and outgoing) from the test graph and construct its feature vector following the encoding scheme described in Section~\ref{sec:subgraph_generation}.

If a node with an identical feature vector exists in the model graph, its embedding is directly used to initialize the entity representation. Otherwise, we identify the most similar node by computing the Hamming distance between feature vectors and select the closest match. The embedding of this nearest node is then used as the initial representation. This procedure enables generalization to unseen entities without requiring retraining.

\paragraph{Evaluation metrics.}
We report two standard ranking-based metrics: Mean Reciprocal Rank (MRR) and Hits@10 (H@10). Hits@10 measures the proportion of test queries for which the correct entity is ranked within the top-10 predictions, while MRR captures the average reciprocal rank of the correct entity.

\paragraph{Ranking protocol.}
For each test triple $(h, r, t)$, we perform both head prediction $(?, r, t)$ and tail prediction $(h, r, ?)$ by ranking the correct entity against 50 randomly sampled negative entities. This sampled evaluation protocol is adopted for computational efficiency.

To ensure fair evaluation, we apply a filtered setting within the sampled candidates by removing any negative samples that correspond to known true triples in the training, validation, or test sets.

This sampling procedure is repeated five times with different random seeds. For each run, results are averaged over head and tail prediction tasks, and the final reported performance is obtained by averaging across all runs.

\subsection{Baselines}

We compare \textsc{MGIL} against representative inductive knowledge graph reasoning methods, categorized into three groups: rule-learning-based methods, subgraph-based GNN methods, and neural message-passing approaches.

\paragraph{Rule-learning-based methods.}
This category includes RuleN~\citep{meilicke2018fine}, which explicitly extracts logical rules from knowledge graphs. 
In addition, Neural-LP~\citep{yang2017differentiable} and DRUM~\citep{sadeghian2019drum} learn logical rules in an end-to-end differentiable framework, enabling gradient-based optimization of rule structures.

\paragraph{Subgraph-based GNN methods.}
GraIL~\citep{teru2020inductiverelationpredictionsubgraph} performs inductive reasoning by extracting enclosing subgraphs around target links and applying graph neural networks for relation prediction. 
CoMPILE~\citep{mai2021communicative} improves upon GraIL by introducing communicative message passing to enhance subgraph representation learning. 
MorsE~\citep{chen2022metaknowledgetransferinductiveknowledge} further extends this line of work by leveraging meta-learning (MAML) to adapt across task-specific subgraphs.

\paragraph{Neural message-passing methods.}
This category includes approaches based on relational graph neural networks, such as R-GCN~\cite{schlichtkrull2018modeling}, which perform message passing over relational graphs to learn entity representations through multi-hop aggregation. 
We additionally compare with NBFNet~\citep{zhu2021neural}, which formulates relational reasoning as a generalized neural Bellman-Ford message passing framework. 
By explicitly modeling path-based dependencies, NBFNet effectively captures multi-hop relational patterns and has demonstrated strong performance on link prediction benchmarks.

\paragraph{Evaluation protocol.}
For all baselines, we report results as published in their original papers, following the corresponding inductive evaluation protocols.
While the specific evaluation settings (e.g., number of negative samples) may vary across methods, all results are based on the Hits@10 metric for link prediction on the same benchmark datasets, enabling meaningful comparison of inductive reasoning performance.

\subsection{Result Analysis}

We evaluate \textsc{MGIL} under two model graph configurations and two experimental settings. 
For the \textbf{relation-based model graph} (denoted as \textsc{MGIL}$_r$), which does not require entity type information, we conduct experiments on: 
(i) standard inductive benchmarks widely adopted in prior literature, and 
(ii) the more challenging inductive splits recently proposed by \citet{Shomer_2025}. 
For the \textbf{entity-based model graph} (denoted as \textsc{MGIL}$_e$), which requires entity type information, we evaluate only on the Hetionet dataset from \citet{Shomer_2025}, as it provides sufficient entity type annotations necessary for this configuration.
To avoid conflating evaluation protocols, we analyze these settings independently.

Table~\ref{tab:results} reports performance on standard inductive benchmarks using \textsc{MGIL}$_r$, including WN18RR-V1--V4, FB15k-237-V1--V4, and NELL-995-V1--V4. 
These datasets follow conventional entity-disjoint inductive splits established in prior work.

Under this evaluation protocol, \textsc{MGIL}$_r$ consistently achieves competitive or superior performance compared to existing baselines across all variants. 
Unlike rule-based methods that rely on explicit structural heuristics, \textsc{MGIL}$_r$ learns transferable structural representations directly from graph topology through its meta-learning framework, enabling effective generalization to unseen entities.

Across different dataset families and split versions (V1--V4), \textsc{MGIL}$_r$ maintains consistent performance, demonstrating robustness to variations in graph density and structural patterns. 
This consistency suggests that the framework effectively captures inductive structural signals that generalize to unseen entities without dataset-specific adaptations.

\begin{table}[h]
	\centering
	 \tiny
	\setlength{\tabcolsep}{3pt}
	\caption{Hits@10 (\%) of inductive link prediction on three datasets. 	Results of baselines are taken from  \cite{chen2022metaknowledgetransferinductiveknowledge}.Bold numbers indicate the highest score in each column }
	
	\begin{tabular}{lcccccccccccc}
		\hline
		\\
		& \multicolumn{4}{c}{WN18RR} & \multicolumn{4}{c}{FB15k-237} & \multicolumn{4}{c}{NELL955} \\
		Method & v1 & v2 & v3 & v4 & v1 & v2 & v3 & v4 & v1 & v2 & v3 & v4 \\
		\hline
		Neural-LP & 74.37 & 68.93 & 46.18 & 67.13 & 52.92 & 58.94 & 52.90 & 55.88 & 40.78 & 78.73 & 82.71 & 80.58 \\
		DRUM      & 74.37 & 68.93 & 46.18 & 67.13 & 52.92 & 58.73 & 52.90 & 55.88 & 19.42 & 78.55 & 82.71 & 80.58 \\
		RuleN     & 80.85 & 78.23 & 53.39 & 71.59 & 49.76 & 77.82 & 87.69 & 85.60 & 53.50 & 81.75 & 77.26 & 61.35 \\
		GraIL     & 82.45 & 78.68 & 58.43 & 73.41 & 64.15 & 81.80 & 82.83 & 89.29 & 59.50 & 93.25 & 91.41 & 73.19 \\
		CoMPILE   & 83.60 & 79.82 & 60.69 & 75.49 & 67.64 & 82.98 & 84.67 & 87.44 & 58.38 & {93.87} & 92.77 & 75.19 \\
		MorsE & 83.21 & 80.42 & 71.64 & \textbf{ 78.70 } & 82.59 & {94.92} & \textbf{95.00} & {95.78} & 62.12 & 80.02 &86.06 & 58.12\\

		\hline
			\textsc{MGIL$_r$} & \textbf{84.86} & \textbf{83.90 }& \textbf{77.02} & {77.81} &\textbf{ 87.17} & \textbf{94.98} & 94.62 &\textbf{96.01} & \textbf{84.40} & \textbf{97.37 } & \textbf{98.80} & \textbf{95.48} \\
		\hline
	\end{tabular}
		
	\label{tab:results}
\end{table}

Table~\ref{tab:new_e_ind_results} presents the results on the 
recently introduced benchmarks of \citet{Shomer_2025}, which 
define stricter inference splits (Inference 1 and Inference 2) 
designed to better simulate realistic inductive scenarios. 
These results are obtained using the \textbf{relation-based model graph}.

On \textbf{CoDEx-M}, \textsc{MGIL}$_r$ achieves 70.11, 
significantly outperforming prior supervised baselines such as NBFNet (43.6) and RED-GNN (35.6). 
This result demonstrates the ability of \textsc{MGIL} to generalize structural knowledge under the revised inductive protocol.

On \textbf{WN18RR}, \textsc{MGIL}$_r$ achieves 75.37 in the Inference 1 setting and 
75.91 in the more challenging Inference 2 setting, outperforming all competing methods. 
Notably, the performance remains stable across both settings, indicating strong robustness 
under stricter entity-disjoint conditions.

On \textbf{HetioNet}, \textsc{MGIL}$_r$ achieves the best performance in both inference settings, 
with scores of 86.12 (Inference 1) and 90.84 (Inference 2), surpassing all baseline methods. 
Given the structural complexity and heterogeneity of HetioNet, these results further confirm 
the effectiveness of the proposed framework in handling diverse knowledge graph structures.

\begin{table*}[t]
	\centering
	 \footnotesize
\caption{Inductive link prediction results (Hits@10) on standard benchmarks. Baseline results are reported from~\citet{Shomer_2025}. Best results are in \textbf{bold}. Results are reported as mean $\pm$ standard deviation over 5 random seeds.}

	\begin{tabular}{l|c|cc|cc}
		
		\toprule
		\multirow{1}{*}{{\bf Models}} & \multicolumn{1}{c}{{\bf CoDEx-M}} &\multicolumn{2}{c}{{\bf WN18RR}}  &\multicolumn{2}{c}{{\bf HetioNet}}   \\ 
		
		& Inference 1 & Inference 1 & Inference 2 & Inference 1 & Inference 2 \\
		\midrule
		
		Neural LP & 13.0 ± 17.9 & 37.9 ± 1.4 & 14.8 ± 1.9 & 12.0 ± 16.4 & 10.7 ± 15.0 \\
		NodePiece & 6.8 ± 0.8 & 29.6 ± 0.8 & 4.8 ± 0.6 & 10.2 ± 0.9	&	15.4 ± 0.9 \\
		InGram & 20.1 ± 3.5 & 38.0 ± 2.4 & 8.0 ± 2.9 & 21.9 ± 1.1 & 22.3 ± 2.8\\
		DEq-InGram & 23.8 ± 1.6 & 62.5 ± 0.8 & 19.1 ± 3.1 & 26.5 ± 4.1 & 28.8 ± 3.5 \\
		RED-GNN &  {35.6 ± 2.3} & {72.9 ± 0.4} & {27.7 ± 0.3} & 68.3 ± 3.0 & {85.1 ± 2.7} \\
		NBFNet &  {43.6 ± 0.2} & \textbf{ 75.5 ± 0.2} & {29.4 ± 2.5} & { 72.8 ± 3.8} & {77.2 ± 0.4} \\

			\hline
		MGIL$_r$ & \textbf{70.11 ± 1.7 } & { 75.37 ± 1.2 } & \textbf{75.91 ±  1.4 } &\textbf{ 86.12 ±  0.78 } & \textbf{90.84 ± 1.3 } \\
		\bottomrule
	\end{tabular}
	\label{tab:new_e_ind_results}
\end{table*}

Table~\ref{tab:shomer_results} presents the MRR performance of 
\textsc{MGIL}$_r$ (relation-based configuration) on Shomer et al. 
benchmarks, which evaluate inductive link prediction under 
challenging structural distribution shifts. 

On \textbf{CoDEx-M Inference 1}, \textsc{MGIL}$_r$ achieves an 
MRR of 52.14, substantially outperforming all baselines 
including NBFNet (27.6) and RED-GNN (22.1), indicating robust 
handling of unseen relational patterns. 

On \textbf{WN18RR}, the results show mixed performance across 
inference settings. In Inference 1, \textsc{MGIL}$_r$ achieves 
58.5, which is lower than RED-GNN (67.8) and NBFNet (66.6). 
However, in the more challenging Inference 2 setting, 
\textsc{MGIL}$_r$ achieves 20.31, modestly outperforming NBFNet 
(17.6) and RED-GNN (16.4), demonstrating improved robustness 
under stricter entity-disjoint conditions.

On \textbf{Hetionet}, \textsc{MGIL}$_r$ achieves competitive 
performance in Inference 1 (55.53, slightly above RED-GNN's 
53.8) and the highest MRR in Inference 2 (71.36), substantially 
surpassing RED-GNN (68.6) and NBFNet (45.2). These results 
highlight \textsc{MGIL}'s ability to leverage meta-learned 
structural priors from the model graph for effective 
generalization to structurally distinct target graphs, 
particularly in highly heterogeneous domains.
 generalization to structurally distinct target graphs.
 
\begin{table}[h]
	\centering
	\footnotesize
\caption{Inductive link prediction results (MRR) on Shomer et al. benchmarks. Baseline results are reported from~\citet{Shomer_2025}. Best results are in \textbf{bold}. All results are reported as mean $\pm$ standard deviation over 5 random seeds.}
	\label{tab:shomer_results}
	\begin{tabular}{lccccc}
		\toprule
		\multirow{2}{*}{Models} & CoDEx-M & \multicolumn{2}{c}{WN18RR} & \multicolumn{2}{c}{Hetionet} \\
		\cmidrule(lr){2-2} \cmidrule(lr){3-4} \cmidrule(lr){5-6}
		& Inference 1 & Inference 1 & Inference 2 & Inference 1 & Inference 2 \\
		\midrule
		Neural LP & 9.2 ± 11.2 & 23.9 ± 2.0 & 10.5 ± 1.8 & 7.3 ± 9.9 & 6.7 ± 9.3 \\
		NodePiece & 3.1 ± 0.3 & 20.6 ± 0.8 & 3.3 ± 0.4 & 5.1 ± 0.6 & 7.2 ± 0.6 \\
		InGram & 10.1 ± 1.5 & 15.9 ± 0.7 & 3.8 ± 1.1 & 11.2 ± 0.5 & 11.5 ± 2.1 \\
		DEq-InGram & 11.2 ± 0.6 & 25.4 ± 0.5 & 9.8 ± 1.7 & 16.5 ± 2.9 & 16.9 ± 2.0 \\
		RED-GNN & 22.1 ± 1.8 & 67.8 ± 0.5 & 16.4 ± 1.3 & 53.8 ± 2.5 & 68.6 ± 4.3 \\
		NBFNet & 27.6 ± 0.2 & \textbf{66.6 ± 0.1} & 17.6 ± 0.1 & 39.1 ± 1.2 & 45.2 ± 0.5 \\
		\midrule
		MGIL$_r$ & \textbf{52.14 ± 0.62 } & 58.5 ± 2.0  & \textbf{20.31 ± 2.8 }  & \textbf{55.53  ± 2.1 } & \textbf{71.36 ± 0.98 } \\
		\bottomrule
	\end{tabular}
\end{table}

To evaluate the effectiveness of the entity-based model graph configuration, we conduct additional experiments on \textbf{Hetionet}, which provides entity type annotations required for this setup.
Unlike the relation-based model graph used across all datasets, the entity-based variant constructs meta-relations based on entity types rather than relation types.
Table~\ref{tab:entity_based_mg} presents the results across two inference runs with different random seeds.

\begin{table}[h]
	\centering
	 \footnotesize
	 

\caption{Inductive link prediction results (Hits@10~(\%)) on Shomer et al.  benchmarks across two inference settings. Best results are in \textbf{bold}. All results are reported as mean $\pm$ standard deviation over 5 random seeds.} 
	\label{tab:entity_based_mg}
	\begin{tabular}{lcc}
		\toprule
		&\multicolumn{2}{c}{{\bf HetioNet}}   \\ 
		 \cmidrule(lr){2-3}
		\textbf{Model} & \textbf{Inference 1} & \textbf{Inference 2} \\
		\midrule
		Neural LP & 12.0 ± 16.4 & 10.7 ± 15.0 \\
		NodePiece & 10.2 ± 0.9 & 15.4 ± 0.9 \\
		InGram & 21.9 ± 1.1 & 22.3 ± 2.8 \\
		DEq-InGram & 26.5 ± 4.1 & 28.8 ± 3.5 \\
		RED-GNN & 68.3 ± 3.0 & 85.1 ± 2.7 \\
		NBFNet & 72.8 ± 3.8 & 77.2 ± 0.4 \\
		MorsE & 84.72 ± 1.8 &86.69 ± 1.8 \\
		\midrule
		\textsc{MGIL$_e$}  & \textbf{90.05± 1.2 } & \textbf{90.57 ± 1} \\
		\bottomrule
	\end{tabular}
\end{table}

With the entity-based model graph, \textsc{MGIL} achieves 90.05\% and 90.57\% on the two inference runs, outperforming all baselines including the strongest competitor MorsE.
These results demonstrate that when entity type information is available, the entity-based model graph configuration can effectively capture entity-level patterns and further enhance inductive reasoning performance.
The consistent performance across both inference runs confirms the robustness of this configuration for heterogeneous knowledge graphs with rich entity type annotations.

Overall, the results across both settings demonstrate that \textsc{MGIL} performs reliably under conventional inductive benchmarks and exhibits even stronger advantages under the more challenging splits introduced by Shomer et al. This separation of evaluation protocols ensures a fair and transparent comparison while highlighting the inductive robustness of the proposed approach.
\section{Conclusion}

In this paper, we introduced Model Graph Inductive Learning (MGIL), a novel framework for inductive knowledge graph completion that leverages a higher-level abstraction of the original knowledge graph, referred to as the model graph. By transforming entities or relations into compact and semantically meaningful nodes, the proposed approach captures global structural patterns that are often overlooked by purely local message-passing methods.

Unlike conventional inductive approaches that rely primarily on subgraph sampling or neighborhood aggregation, MGIL enhances representation learning by incorporating structural regularities at the model graph level. In particular, embeddings learned over the model graph provide a meaningful and informed initialization for entity representations, which facilitates more effective and stable GNN-based message passing in subsequent stages. This leads to improved generalization to unseen entities and relations, particularly in sparse or low-resource scenarios.

Extensive experiments on multiple inductive benchmarks, including CoDEx-M E, WN18RR E, and HetioNet E, demonstrate the effectiveness of the proposed framework. The results show that MGIL consistently outperforms strong baselines, highlighting the importance of incorporating global structural information into inductive reasoning.

Furthermore, the flexibility of the model graph construction ,such as relation-based and entity-type-based variants , provides a unified perspective for capturing different aspects of knowledge graphs. Beyond simple abstractions, the model graph can be extended to more expressive forms, such as directed and weighted structures, or even multi-relational graphs, enabling richer representations of dependencies between abstract nodes.

Overall, MGIL provides a principled framework for bridging local and global reasoning in knowledge graph completion.

%
\bibliographystyle{plain}

\begin{thebibliography}{10}
	\ifx \bisbn   \undefined \def \bisbn  #1{ISBN #1}\fi
	\ifx \binits  \undefined \def \binits#1{#1}\fi
	\ifx \bauthor  \undefined \def \bauthor#1{#1}\fi
	\ifx \batitle  \undefined \def \batitle#1{#1}\fi
	\ifx \bjtitle  \undefined \def \bjtitle#1{#1}\fi
	\ifx \bvolume  \undefined \def \bvolume#1{\textbf{#1}}\fi
	\ifx \byear  \undefined \def \byear#1{#1}\fi
	\ifx \bissue  \undefined \def \bissue#1{#1}\fi
	\ifx \bfpage  \undefined \def \bfpage#1{#1}\fi
	\ifx \blpage  \undefined \def \blpage #1{#1}\fi
	\ifx \burl  \undefined \def \burl#1{\textsf{#1}}\fi
	\ifx \doiurl  \undefined \def \doiurl#1{\url{https://doi.org/#1}}\fi
	\ifx \betal  \undefined \def \betal{\textit{et al.}}\fi
	\ifx \binstitute  \undefined \def \binstitute#1{#1}\fi
	\ifx \binstitutionaled  \undefined \def \binstitutionaled#1{#1}\fi
	\ifx \bctitle  \undefined \def \bctitle#1{#1}\fi
	\ifx \beditor  \undefined \def \beditor#1{#1}\fi
	\ifx \bpublisher  \undefined \def \bpublisher#1{#1}\fi
	\ifx \bbtitle  \undefined \def \bbtitle#1{#1}\fi
	\ifx \bedition  \undefined \def \bedition#1{#1}\fi
	\ifx \bseriesno  \undefined \def \bseriesno#1{#1}\fi
	\ifx \blocation  \undefined \def \blocation#1{#1}\fi
	\ifx \bsertitle  \undefined \def \bsertitle#1{#1}\fi
	\ifx \bsnm \undefined \def \bsnm#1{#1}\fi
	\ifx \bsuffix \undefined \def \bsuffix#1{#1}\fi
	\ifx \bparticle \undefined \def \bparticle#1{#1}\fi
	\ifx \barticle \undefined \def \barticle#1{#1}\fi
	\bibcommenthead
	\ifx \bconfdate \undefined \def \bconfdate #1{#1}\fi
	\ifx \botherref \undefined \def \botherref #1{#1}\fi
	\ifx \url \undefined \def \url#1{\textsf{#1}}\fi
	\ifx \bchapter \undefined \def \bchapter#1{#1}\fi
	\ifx \bbook \undefined \def \bbook#1{#1}\fi
	\ifx \bcomment \undefined \def \bcomment#1{#1}\fi
	\ifx \oauthor \undefined \def \oauthor#1{#1}\fi
	\ifx \citeauthoryear \undefined \def \citeauthoryear#1{#1}\fi
	\ifx \endbibitem  \undefined \def \endbibitem {}\fi
	\ifx \bconflocation  \undefined \def \bconflocation#1{#1}\fi
	\ifx \arxivurl  \undefined \def \arxivurl#1{\textsf{#1}}\fi
	\csname PreBibitemsHook\endcsname
	
	\bibitem[\protect\citeauthoryear{Bordes et~al.}{2013}]{bordes2013transe}
	\begin{bchapter}
		\bauthor{\bsnm{Bordes}, \binits{A.}},
		\bauthor{\bsnm{Usunier}, \binits{N.}},
		\bauthor{\bsnm{Garcia-Duran}, \binits{A.}},
		\bauthor{\bsnm{Weston}, \binits{J.}},
		\bauthor{\bsnm{Yakhnenko}, \binits{O.}}:
		\bctitle{Translating embeddings for modeling multi-relational data}.
		In: \bbtitle{Advances in Neural Information Processing Systems (NeurIPS)},
		pp. \bfpage{2787}--\blpage{2795}
		(\byear{2013})
	\end{bchapter}
	\endbibitem
	
	\bibitem[\protect\citeauthoryear{Wang et~al.}{2014}]{wang2014transh}
	\begin{bchapter}
		\bauthor{\bsnm{Wang}, \binits{Z.}},
		\bauthor{\bsnm{Zhang}, \binits{J.}},
		\bauthor{\bsnm{Feng}, \binits{J.}},
		\bauthor{\bsnm{Chen}, \binits{Z.}}:
		\bctitle{Knowledge graph embedding by translating on hyperplanes}.
		In: \bbtitle{Proceedings of the Twenty-Eighth AAAI Conference on Artificial
			Intelligence (AAAI)},
		pp. \bfpage{1112}--\blpage{1119}
		(\byear{2014})
	\end{bchapter}
	\endbibitem
	
	\bibitem[\protect\citeauthoryear{Yang et~al.}{2015}]{yang2015embedding}
	\begin{bchapter}
		\bauthor{\bsnm{Yang}, \binits{B.}},
		\bauthor{\bsnm{Yih}, \binits{W.-t.}},
		\bauthor{\bsnm{He}, \binits{X.}},
		\bauthor{\bsnm{Gao}, \binits{J.}},
		\bauthor{\bsnm{Deng}, \binits{L.}}:
		\bctitle{Embedding entities and relations for learning and inference in
			knowledge bases}.
		In: \beditor{\bsnm{Bengio}, \binits{Y.}},
		\beditor{\bsnm{LeCun}, \binits{Y.}} (eds.)
		\bbtitle{Proceedings of the International Conference on Learning
			Representations (ICLR)}
		(\byear{2015})
	\end{bchapter}
	\endbibitem
	
	\bibitem[\protect\citeauthoryear{Wang et~al.}{2019}]{wang2019knowledge}
	\begin{bchapter}
		\bauthor{\bsnm{Wang}, \binits{H.}},
		\bauthor{\bsnm{Zhao}, \binits{M.}},
		\bauthor{\bsnm{Xie}, \binits{X.}},
		\bauthor{\bsnm{Li}, \binits{W.}},
		\bauthor{\bsnm{Guo}, \binits{M.}}:
		\bctitle{Knowledge graph convolutional networks for recommender systems}.
		In: \bbtitle{Proceedings of the 25th ACM SIGKDD International Conference on
			Knowledge Discovery \& Data Mining},
		pp. \bfpage{950}--\blpage{958}
		(\byear{2019})
	\end{bchapter}
	\endbibitem
	
	\bibitem[\protect\citeauthoryear{Cao et~al.}{2019}]{cao2019unifying}
	\begin{bchapter}
		\bauthor{\bsnm{Cao}, \binits{Y.}},
		\bauthor{\bsnm{Wang}, \binits{X.}},
		\bauthor{\bsnm{He}, \binits{X.}},
		\bauthor{\bsnm{Hu}, \binits{Z.}},
		\bauthor{\bsnm{Chua}, \binits{T.-S.}}:
		\bctitle{Unifying knowledge graph learning and recommendation: Towards a better
			understanding of user preferences}.
		In: \bbtitle{2019 World Wide Web Conference (WWW)},
		pp. \bfpage{151}--\blpage{161}
		(\byear{2019})
	\end{bchapter}
	\endbibitem
	
	\bibitem[\protect\citeauthoryear{Wang et~al.}{2017}]{8047276}
	\begin{barticle}
		\bauthor{\bsnm{Wang}, \binits{Q.}},
		\bauthor{\bsnm{Mao}, \binits{Z.}},
		\bauthor{\bsnm{Wang}, \binits{B.}},
		\bauthor{\bsnm{Guo}, \binits{L.}}:
		\batitle{{ Knowledge Graph Embedding: A Survey of Approaches and Applications
		}}.
		\bjtitle{IEEE Transactions on Knowledge \& Data Engineering}
		\bvolume{29}(\bissue{12}),
		\bfpage{2724}--\blpage{2743}
		(\byear{2017})
	\end{barticle}
	\endbibitem
	
	\bibitem[\protect\citeauthoryear{Boroujeni et~al.}{2022}]{boroujeni2022answer}
	\begin{barticle}
		\bauthor{\bsnm{Boroujeni}, \binits{G.A.}},
		\bauthor{\bsnm{Faili}, \binits{H.}},
		\bauthor{\bsnm{Yaghoobzadeh}, \binits{Y.}}:
		\batitle{Answer selection in community question answering exploiting knowledge
			graph and context information}.
		\bjtitle{Semantic Web}
		\bvolume{13}(\bissue{3}),
		\bfpage{339}--\blpage{356}
		(\byear{2022})
	\end{barticle}
	\endbibitem
	
	\bibitem[\protect\citeauthoryear{Trouillon et~al.}{2016}]{trouillon2016complex}
	\begin{bchapter}
		\bauthor{\bsnm{Trouillon}, \binits{T.}},
		\bauthor{\bsnm{Welbl}, \binits{J.}},
		\bauthor{\bsnm{Riedel}, \binits{S.}},
		\bauthor{\bsnm{Gaussier}, \binits{{\'E}.}},
		\bauthor{\bsnm{Bouchard}, \binits{G.}}:
		\bctitle{Complex embeddings for simple link prediction}.
		In: \bbtitle{Proceedings of the 33rd International Conference on Machine
			Learning (ICML)}.
		\bsertitle{Proceedings of Machine Learning Research},
		vol. \bseriesno{48},
		pp. \bfpage{2071}--\blpage{2080}.
		\bpublisher{PMLR},
		\blocation{New York, NY, USA}
		(\byear{2016})
	\end{bchapter}
	\endbibitem
	
	\bibitem[\protect\citeauthoryear{Sun et~al.}{2019}]{sun2019rotate}
	\begin{bchapter}
		\bauthor{\bsnm{Sun}, \binits{Z.}},
		\bauthor{\bsnm{Deng}, \binits{Z.-H.}},
		\bauthor{\bsnm{Nie}, \binits{J.-Y.}},
		\bauthor{\bsnm{Tang}, \binits{J.}}:
		\bctitle{Rotate: Knowledge graph embedding by relational rotation in complex
			space}.
		In: \bbtitle{International Conference on Learning Representations}
		(\byear{2019})
	\end{bchapter}
	\endbibitem
	
	\bibitem[\protect\citeauthoryear{Shi and Weninger}{2018}]{shi2018open}
	\begin{botherref}
		\oauthor{\bsnm{Shi}, \binits{B.}},
		\oauthor{\bsnm{Weninger}, \binits{T.}}:
		Open-world knowledge graph completion.
		Proceedings of the AAAI Conference on Artificial Intelligence
		\textbf{32}(1)
		(2018)
	\end{botherref}
	\endbibitem
	
	\bibitem[\protect\citeauthoryear{Teru
		et~al.}{2020}]{teru2020inductiverelationpredictionsubgraph}
	\begin{bchapter}
		\bauthor{\bsnm{Teru}, \binits{K.}},
		\bauthor{\bsnm{Denis}, \binits{E.}},
		\bauthor{\bsnm{Hamilton}, \binits{W.}}:
		\bctitle{Inductive relation prediction by subgraph reasoning}.
		In: \beditor{\bsnm{III}, \binits{H.D.}},
		\beditor{\bsnm{Singh}, \binits{A.}} (eds.)
		\bbtitle{Proceedings of the 37th International Conference on Machine Learning}.
		\bsertitle{Proceedings of Machine Learning Research},
		vol. \bseriesno{119},
		pp. \bfpage{9448}--\blpage{9457}.
		\bpublisher{PMLR}, \blocation{???}
		(\byear{2020})
	\end{bchapter}
	\endbibitem
	
	\bibitem[\protect\citeauthoryear{Xu et~al.}{}]{xu2022subgraph}
	\begin{botherref}
		\oauthor{\bsnm{Xu}, \binits{X.}},
		\oauthor{\bsnm{Zhang}, \binits{P.}},
		\oauthor{\bsnm{He}, \binits{Y.}},
		\oauthor{\bsnm{Chao}, \binits{C.}},
		\oauthor{\bsnm{Yan}, \binits{C.}}:
		Subgraph neighboring relations infomax for inductive link prediction on
		knowledge graphs.
		arXiv preprint arXiv:2208.00850,
		2316--2322
	\end{botherref}
	\endbibitem
	
	\bibitem[\protect\citeauthoryear{Katz}{1953}]{katz1953new}
	\begin{barticle}
		\bauthor{\bsnm{Katz}, \binits{L.}}:
		\batitle{A new status index derived from sociometric analysis}.
		\bjtitle{Psychometrika}
		\bvolume{18}(\bissue{1}),
		\bfpage{39}--\blpage{43}
		(\byear{1953})
	\end{barticle}
	\endbibitem
	
	\bibitem[\protect\citeauthoryear{Page et~al.}{1999}]{page1999pagerank}
	\begin{bchapter}
		\bauthor{\bsnm{Page}, \binits{L.}},
		\bauthor{\bsnm{Brin}, \binits{S.}},
		\bauthor{\bsnm{Motwani}, \binits{R.}},
		\bauthor{\bsnm{Winograd}, \binits{T.}}:
		\bctitle{The pagerank citation ranking: Bring order to the web}.
		In: \bbtitle{Proc. of the 7th International World Wide Web Conf.--1998}
		(\byear{1999})
	\end{bchapter}
	\endbibitem
	
	\bibitem[\protect\citeauthoryear{Dijkstra}{2022}]{dijkstra1959note}
	\begin{bchapter}
		\bauthor{\bsnm{Dijkstra}, \binits{E.W.}}:
		\bctitle{A note on two problems in connexion with graphs}.
		In: \bbtitle{Edsger Wybe Dijkstra: His Life, Work, and Legacy},
		pp. \bfpage{287}--\blpage{290}
		(\byear{2022})
	\end{bchapter}
	\endbibitem
	
	\bibitem[\protect\citeauthoryear{Jeh and Widom}{2002}]{jeh2002simrank}
	\begin{bchapter}
		\bauthor{\bsnm{Jeh}, \binits{G.}},
		\bauthor{\bsnm{Widom}, \binits{J.}}:
		\bctitle{Simrank: a measure of structural-context similarity}.
		In: \bbtitle{Proceedings of the Eighth ACM SIGKDD International Conference on
			Knowledge Discovery and Data Mining (KDD 2002)},
		pp. \bfpage{538}--\blpage{543}.
		\bpublisher{ACM},
		\blocation{~}
		(\byear{2002})
	\end{bchapter}
	\endbibitem
	
	\bibitem[\protect\citeauthoryear{Sun et~al.}{2011}]{sun2011pathsim}
	\begin{botherref}
		\oauthor{\bsnm{Sun}, \binits{Y.}},
		\oauthor{\bsnm{Han}, \binits{J.}},
		\oauthor{\bsnm{Yan}, \binits{X.}},
		\oauthor{\bsnm{Yu}, \binits{P.S.}},
		\oauthor{\bsnm{Wu}, \binits{T.}}:
		Pathsim: meta path-based top-k similarity search in heterogeneous information
		networks.
		Proc. VLDB Endow.
		\textbf{4}(11)
		(2011)
	\end{botherref}
	\endbibitem
	
	\bibitem[\protect\citeauthoryear{Lao and Cohen}{2010}]{lao2010relational}
	\begin{barticle}
		\bauthor{\bsnm{Lao}, \binits{N.}},
		\bauthor{\bsnm{Cohen}, \binits{W.W.}}:
		\batitle{Relational retrieval using a combination of path-constrained random
			walks}.
		\bjtitle{Machine Learning}
		\bvolume{81}(\bissue{1}),
		\bfpage{53}--\blpage{67}
		(\byear{2010})
	\end{barticle}
	\endbibitem
	
	\bibitem[\protect\citeauthoryear{Neelakantan
		et~al.}{2015}]{neelakantan2015compositional}
	\begin{bchapter}
		\bauthor{\bsnm{Neelakantan}, \binits{A.}},
		\bauthor{\bsnm{Roth}, \binits{B.}},
		\bauthor{\bsnm{McCallum}, \binits{A.}}:
		\bctitle{Compositional vector space models for knowledge base completion}.
		In: \bbtitle{Proceedings of the 53rd Annual Meeting of the Association for
			Computational Linguistics and the 7th International Joint Conference on
			Natural Language Processing (Volume 1: Long Papers)},
		pp. \bfpage{156}--\blpage{166}
		(\byear{2015})
	\end{bchapter}
	\endbibitem
	
	\bibitem[\protect\citeauthoryear{Gal{\'a}rraga
		et~al.}{2013}]{galarraga2013amie}
	\begin{bchapter}
		\bauthor{\bsnm{Gal{\'a}rraga}, \binits{L.A.}},
		\bauthor{\bsnm{Teflioudi}, \binits{C.}},
		\bauthor{\bsnm{Hose}, \binits{K.}},
		\bauthor{\bsnm{Suchanek}, \binits{F.}}:
		\bctitle{Amie: Association rule mining under incomplete evidence in ontological
			knowledge bases}.
		In: \bbtitle{Proceedings of the 22nd International Conference on World Wide Web
			(WWW 2013)},
		pp. \bfpage{413}--\blpage{422}.
		\bpublisher{ACM},
		\blocation{~}
		(\byear{2013})
	\end{bchapter}
	\endbibitem
	
	\bibitem[\protect\citeauthoryear{Meilicke et~al.}{2018}]{meilicke2018fine}
	\begin{bchapter}
		\bauthor{\bsnm{Meilicke}, \binits{C.}},
		\bauthor{\bsnm{Fink}, \binits{M.}},
		\bauthor{\bsnm{Wang}, \binits{Y.}},
		\bauthor{\bsnm{Ruffinelli}, \binits{D.}},
		\bauthor{\bsnm{Gemulla}, \binits{R.}},
		\bauthor{\bsnm{Stuckenschmidt}, \binits{H.}}:
		\bctitle{Fine-grained evaluation of rule- and embedding-based systems for
			knowledge graph completion}.
		In: \beditor{\bsnm{Vrande{\v{c}}i{\'{c}}}, \binits{D.}},
		\beditor{\bsnm{Bontcheva}, \binits{K.}},
		\beditor{\bsnm{Su{\'a}rez-Figueroa}, \binits{M.C.}},
		\beditor{\bsnm{Presutti}, \binits{V.}},
		\beditor{\bsnm{Celino}, \binits{I.}},
		\beditor{\bsnm{Sabou}, \binits{M.}},
		\beditor{\bsnm{Kaffee}, \binits{L.-A.}},
		\beditor{\bsnm{Simperl}, \binits{E.}} (eds.)
		\bbtitle{The Semantic Web -- ISWC 2018},
		pp. \bfpage{3}--\blpage{20}.
		\bpublisher{Springer},
		\blocation{Cham}
		(\byear{2018})
	\end{bchapter}
	\endbibitem
	
	\bibitem[\protect\citeauthoryear{Yang et~al.}{2017}]{yang2017differentiable}
	\begin{bchapter}
		\bauthor{\bsnm{Yang}, \binits{F.}},
		\bauthor{\bsnm{Yang}, \binits{Z.}},
		\bauthor{\bsnm{Cohen}, \binits{W.W.}}:
		\bctitle{Differentiable learning of logical rules for knowledge base
			reasoning}.
		In: \beditor{\bsnm{Guyon}, \binits{I.}},
		\beditor{\bsnm{Luxburg}, \binits{U.V.}},
		\beditor{\bsnm{Bengio}, \binits{S.}},
		\beditor{\bsnm{Wallach}, \binits{H.}},
		\beditor{\bsnm{Fergus}, \binits{R.}},
		\beditor{\bsnm{Vishwanathan}, \binits{S.}},
		\beditor{\bsnm{Garnett}, \binits{R.}} (eds.)
		\bbtitle{Advances in Neural Information Processing Systems},
		vol. \bseriesno{30}.
		\bpublisher{Curran Associates, Inc.},
		\blocation{~}
		(\byear{2017})
	\end{bchapter}
	\endbibitem
	
	\bibitem[\protect\citeauthoryear{Sadeghian et~al.}{2019}]{sadeghian2019drum}
	\begin{bchapter}
		\bauthor{\bsnm{Sadeghian}, \binits{A.}},
		\bauthor{\bsnm{Armandpour}, \binits{M.}},
		\bauthor{\bsnm{Ding}, \binits{P.}},
		\bauthor{\bsnm{Wang}, \binits{D.Z.}}:
		\bctitle{Drum: End-to-end differentiable rule mining on knowledge graphs}.
		In: \beditor{\bsnm{Wallach}, \binits{H.}},
		\beditor{\bsnm{Larochelle}, \binits{H.}},
		\beditor{\bsnm{Beygelzimer}, \binits{A.}},
		\beditor{\bsnm{Alch\'{e}-Buc}, \binits{F.}},
		\beditor{\bsnm{Fox}, \binits{E.}},
		\beditor{\bsnm{Garnett}, \binits{R.}} (eds.)
		\bbtitle{Advances in Neural Information Processing Systems},
		vol. \bseriesno{32}.
		\bpublisher{Curran Associates, Inc.},
		\blocation{~}
		(\byear{2019})
	\end{bchapter}
	\endbibitem
	
	\bibitem[\protect\citeauthoryear{Schlichtkrull
		et~al.}{2018}]{schlichtkrull2018modeling}
	\begin{bchapter}
		\bauthor{\bsnm{Schlichtkrull}, \binits{M.}},
		\bauthor{\bsnm{Kipf}, \binits{T.N.}},
		\bauthor{\bsnm{Bloem}, \binits{P.}},
		\bauthor{\bsnm{Berg}, \binits{R.}},
		\bauthor{\bsnm{Titov}, \binits{I.}},
		\bauthor{\bsnm{Welling}, \binits{M.}}:
		\bctitle{Modeling relational data with graph convolutional networks}.
		In: \bbtitle{Proceedings of the Semantic Web (ESWC)},
		pp. \bfpage{593}--\blpage{607}.
		\bpublisher{Springer},
		\blocation{Cham}
		(\byear{2018})
	\end{bchapter}
	\endbibitem
	
	\bibitem[\protect\citeauthoryear{Vashishth
		et~al.}{2020}]{vashishth2020composition}
	\begin{botherref}
		\oauthor{\bsnm{Vashishth}, \binits{S.}},
		\oauthor{\bsnm{Sanyal}, \binits{S.}},
		\oauthor{\bsnm{Nitin}, \binits{V.}},
		\oauthor{\bsnm{Talukdar}, \binits{P.}}:
		Composition-based multi-relational graph convolutional networks
		(2020)
	\end{botherref}
	\endbibitem
	
	\bibitem[\protect\citeauthoryear{Mai et~al.}{2021}]{mai2021communicative}
	\begin{bchapter}
		\bauthor{\bsnm{Mai}, \binits{S.}},
		\bauthor{\bsnm{Zheng}, \binits{S.}},
		\bauthor{\bsnm{Yang}, \binits{Y.}},
		\bauthor{\bsnm{Hu}, \binits{H.}}:
		\bctitle{Communicative message passing for inductive relation reasoning},
		vol. \bseriesno{35},
		pp. \bfpage{4294}--\blpage{4302}
		(\byear{2021})
	\end{bchapter}
	\endbibitem
	
	\bibitem[\protect\citeauthoryear{Yan et~al.}{2022}]{an2022cycle}
	\begin{bchapter}
		\bauthor{\bsnm{Yan}, \binits{Z.}},
		\bauthor{\bsnm{Ma}, \binits{T.}},
		\bauthor{\bsnm{Gao}, \binits{L.}},
		\bauthor{\bsnm{Tang}, \binits{Z.}},
		\bauthor{\bsnm{Chen}, \binits{C.}}:
		\bctitle{Cycle representation learning for inductive relation prediction}.
		In: \beditor{\bsnm{Chaudhuri}, \binits{K.}},
		\beditor{\bsnm{Jegelka}, \binits{S.}},
		\beditor{\bsnm{Song}, \binits{L.}},
		\beditor{\bsnm{Szepesvari}, \binits{C.}},
		\beditor{\bsnm{Niu}, \binits{G.}},
		\beditor{\bsnm{Sabato}, \binits{S.}} (eds.)
		\bbtitle{Proceedings of the 39th International Conference on Machine Learning}.
		\bsertitle{Proceedings of Machine Learning Research},
		vol. \bseriesno{162},
		pp. \bfpage{24895}--\blpage{24910}.
		\bpublisher{PMLR},
		\blocation{~}
		(\byear{2022})
	\end{bchapter}
	\endbibitem
	
	\bibitem[\protect\citeauthoryear{Shomer et~al.}{2025}]{Shomer_2025}
	\begin{bchapter}
		\bauthor{\bsnm{Shomer}, \binits{H.}},
		\bauthor{\bsnm{Revolinsky}, \binits{J.}},
		\bauthor{\bsnm{Tang}, \binits{J.}}:
		\bctitle{Towards better benchmark datasets for inductive knowledge graph
			completion}.
		In: \beditor{\bsnm{Antonie}, \binits{L.}},
		\beditor{\bsnm{Pei}, \binits{J.}},
		\beditor{\bsnm{Yu}, \binits{X.}},
		\beditor{\bsnm{Chierichetti}, \binits{F.}},
		\beditor{\bsnm{Lauw}, \binits{H.W.}},
		\beditor{\bsnm{Sun}, \binits{Y.}},
		\beditor{\bsnm{Parthasarathy}, \binits{S.}} (eds.)
		\bbtitle{Proceedings of the 31st {ACM} {SIGKDD} Conference on Knowledge
			Discovery and Data Mining, V.2, {KDD} 2025, Toronto ON, Canada, August 3-7,
			2025}.
		\bsertitle{KDD ’25},
		pp. \bfpage{5777}--\blpage{5788}.
		\bpublisher{{ACM}},
		\blocation{~}
		(\byear{2025})
	\end{bchapter}
	\endbibitem
	
	\bibitem[\protect\citeauthoryear{Chen
		et~al.}{2022}]{chen2022metaknowledgetransferinductiveknowledge}
	\begin{bchapter}
		\bauthor{\bsnm{Chen}, \binits{M.}},
		\bauthor{\bsnm{Zhang}, \binits{W.}},
		\bauthor{\bsnm{Zhu}, \binits{Y.}},
		\bauthor{\bsnm{Zhou}, \binits{H.}},
		\bauthor{\bsnm{Yuan}, \binits{Z.}},
		\bauthor{\bsnm{Xu}, \binits{C.}},
		\bauthor{\bsnm{Chen}, \binits{H.}}:
		\bctitle{Meta-knowledge transfer for inductive knowledge graph embedding}.
		In: \bbtitle{Proceedings of the 45th International ACM SIGIR Conference on
			Research and Development in Information Retrieval}.
		\bsertitle{SIGIR '22},
		pp. \bfpage{927}--\blpage{937}.
		\bpublisher{Association for Computing Machinery},
		\blocation{New York, NY, USA}
		(\byear{2022})
	\end{bchapter}
	\endbibitem
	
	\bibitem[\protect\citeauthoryear{Paszke et~al.}{2019}]{paszke2019pytorch}
	\begin{bchapter}
		\bauthor{\bsnm{Paszke}, \binits{A.}},
		\bauthor{\bsnm{Gross}, \binits{S.}},
		\bauthor{\bsnm{Massa}, \binits{F.}},
		\bauthor{\bsnm{Lerer}, \binits{A.}},
		\bauthor{\bsnm{Bradbury}, \binits{J.}},
		\bauthor{\bsnm{Chanan}, \binits{G.}},
		\bauthor{\bsnm{Killeen}, \binits{T.}},
		\bauthor{\bsnm{Lin}, \binits{Z.}},
		\bauthor{\bsnm{Gimelshein}, \binits{N.}},
		\bauthor{\bsnm{Antiga}, \binits{L.}},
		\bauthor{\bsnm{Desmaison}, \binits{A.}},
		\bauthor{\bsnm{Kopf}, \binits{A.}},
		\bauthor{\bsnm{Yang}, \binits{E.}},
		\bauthor{\bsnm{DeVito}, \binits{Z.}},
		\bauthor{\bsnm{Raison}, \binits{M.}},
		\bauthor{\bsnm{Tejani}, \binits{A.}},
		\bauthor{\bsnm{Chilamkurthy}, \binits{S.}},
		\bauthor{\bsnm{Steiner}, \binits{B.}},
		\bauthor{\bsnm{Fang}, \binits{L.}},
		\bauthor{\bsnm{Bai}, \binits{J.}},
		\bauthor{\bsnm{Chintala}, \binits{S.}}:
		\bctitle{Pytorch: An imperative style, high-performance deep learning library}.
		In: \bbtitle{Advances in Neural Information Processing Systems (NeurIPS 2019)},
		pp. \bfpage{8024}--\blpage{8035}
		(\byear{2019})
	\end{bchapter}
	\endbibitem
	
	\bibitem[\protect\citeauthoryear{Wang et~al.}{2019}]{wang2019dgl}
	\begin{botherref}
		\oauthor{\bsnm{Wang}, \binits{M.}},
		\oauthor{\bsnm{Zheng}, \binits{D.}},
		\oauthor{\bsnm{Ye}, \binits{Z.}},
		\oauthor{\bsnm{Gan}, \binits{Q.}},
		\oauthor{\bsnm{Li}, \binits{M.}},
		\oauthor{\bsnm{Song}, \binits{X.}},
		\oauthor{\bsnm{Zhou}, \binits{J.}},
		\oauthor{\bsnm{Ma}, \binits{C.}},
		\oauthor{\bsnm{Yu}, \binits{L.}},
		\oauthor{\bsnm{Gai}, \binits{Y.}},
		\oauthor{\bsnm{Xiao}, \binits{T.}},
		\oauthor{\bsnm{He}, \binits{T.}},
		\oauthor{\bsnm{Karypis}, \binits{G.}},
		\oauthor{\bsnm{Li}, \binits{J.}},
		\oauthor{\bsnm{Zhang}, \binits{Z.}}:
		Deep graph library: A graph-centric, highly-performant package for graph neural
		networks.
		arXiv preprint arXiv:1909.01315
		(2019)
	\end{botherref}
	\endbibitem
	
	\bibitem[\protect\citeauthoryear{Dettmers et~al.}{2018}]{dettmers2018conve}
	\begin{bchapter}
		\bauthor{\bsnm{Dettmers}, \binits{T.}},
		\bauthor{\bsnm{Minervini}, \binits{P.}},
		\bauthor{\bsnm{Stenetorp}, \binits{P.}},
		\bauthor{\bsnm{Riedel}, \binits{S.}}:
		\bctitle{Convolutional {2D} knowledge graph embeddings}.
		In: \bbtitle{Proceedings of the 32nd AAAI Conference on Artificial Intelligence
			(AAAI)},
		pp. \bfpage{1811}--\blpage{1818}
		(\byear{2018})
	\end{bchapter}
	\endbibitem
	
	\bibitem[\protect\citeauthoryear{Toutanova and
		Chen}{2015}]{toutanova2015observed}
	\begin{bchapter}
		\bauthor{\bsnm{Toutanova}, \binits{K.}},
		\bauthor{\bsnm{Chen}, \binits{D.}}:
		\bctitle{Observed versus latent features for knowledge base and text
			inference}.
		In: \beditor{\bsnm{Allauzen}, \binits{A.}},
		\beditor{\bsnm{Grefenstette}, \binits{E.}},
		\beditor{\bsnm{Hermann}, \binits{K.M.}},
		\beditor{\bsnm{Larochelle}, \binits{H.}},
		\beditor{\bsnm{Yih}, \binits{S.W.-t.}} (eds.)
		\bbtitle{Proceedings of the 3rd Workshop on Continuous Vector Space Models and
			Their Compositionality},
		pp. \bfpage{57}--\blpage{66}.
		\bpublisher{Association for Computational Linguistics},
		\blocation{Beijing, China}
		(\byear{2015})
	\end{bchapter}
	\endbibitem
	
	\bibitem[\protect\citeauthoryear{Xiong et~al.}{2017}]{xiong2017deeppath}
	\begin{bchapter}
		\bauthor{\bsnm{Xiong}, \binits{W.}},
		\bauthor{\bsnm{Hoang}, \binits{T.}},
		\bauthor{\bsnm{Wang}, \binits{W.Y.}}:
		\bctitle{Deeppath: A reinforcement learning method for knowledge graph
			reasoning}.
		In: \beditor{\bsnm{Palmer}, \binits{M.}},
		\beditor{\bsnm{Hwa}, \binits{R.}},
		\beditor{\bsnm{Riedel}, \binits{S.}} (eds.)
		\bbtitle{Proceedings of the 2017 Conference on Empirical Methods in Natural
			Language Processing},
		pp. \bfpage{564}--\blpage{573}.
		\bpublisher{Association for Computational Linguistics},
		\blocation{Copenhagen, Denmark}
		(\byear{2017})
	\end{bchapter}
	\endbibitem
	
	\bibitem[\protect\citeauthoryear{Zhu et~al.}{2021}]{zhu2021neural}
	\begin{bchapter}
		\bauthor{\bsnm{Zhu}, \binits{Z.}},
		\bauthor{\bsnm{Zhang}, \binits{Z.}},
		\bauthor{\bsnm{Xhonneux}, \binits{L.}},
		\bauthor{\bsnm{Tang}, \binits{J.}}:
		\bctitle{Neural bellman-ford networks: A general graph neural network framework
			for link prediction}.
		In: \bbtitle{Advances in Neural Information Processing Systems (NeurIPS)}.
		\bsertitle{NIPS\'21}.
		\bpublisher{Curran Associates Inc.},
		\blocation{~}
		(\byear{2021})
	\end{bchapter}
	\endbibitem
	
\end{thebibliography}

\appendix

\section{Experimental and Implementation Details}
\label{sec:appendix_implementation}

To ensure the reproducibility and transparency of our experiments, we provide detailed descriptions of the implementation, datasets, baseline configurations, and training procedures.

\subsection{Hardware and Frameworks}
All experiments are conducted on a single NVIDIA RTX 3090 GPU with 24GB of VRAM. 
Our models are implemented using PyTorch~\cite{paszke2019pytorch} and DGL~\cite{wang2019dgl}.

\subsection{Baseline Configurations and Fair Comparison}
We follow a strict protocol to ensure fair and meaningful comparisons with prior work. 
In particular, for baseline models such as \textbf{MORSE}, we adhere to the original experimental settings, evaluation protocols, and hyperparameter configurations reported in their respective papers.

To avoid any potential data leakage, all evaluations are conducted on the  official dataset splits provided by each benchmark, without any modification or subgraph resampling. 

\subsection{Model Architecture and Hyperparameters}
Our framework consists of two main components: a model graph encoder and a task-specific reasoning module.

First, we apply a 2-layer GNN on the model graph to obtain node embeddings of dimension $d=32$. 
These embeddings are then used to parameterize relation-specific transformations in a downstream R-GCN applied to the task graph.

Depending on the dataset, we use either a 2-layer or 3-layer R-GCN, with all hidden and embedding dimensions fixed to 32. 
To regularize relation-specific parameters and reduce model complexity, we adopt the basis decomposition technique proposed by Schlichtkrull et al.~\cite{schlichtkrull2018modeling}, using $B \in \{3,4\}$ bases depending on the number of relations.

To ensure a well-grounded experimental setup, our hyperparameter choices are informed by prior state-of-the-art studies, particularly MORSE and NBFNet, which demonstrate strong empirical performance under comparable hardware and evaluation settings. 
We use their reported configurations and publicly available implementations as a reference point, and introduce only minimal modifications where necessary to align with our model design.

The training configuration is as follows: batch size of 64 tasks, learning rate of 0.01, and 10 training epochs. 
We use a fixed margin $\gamma=10$ and uniformly sample $\beta=32$ negative entities per positive triple.

\subsection{Task Generation}
We generate training and validation tasks using a random-walk-based procedure. 
For each dataset, we construct 10,000 training tasks and 200 validation tasks.

The random walk parameters are set to walk length $l_{rw}=5$, number of walks per entity $n_{rw}=10$, and anonymization ratio $t_{rw}=10$. 
This process ensures that task graphs capture local structural patterns while maintaining diversity across tasks.

\subsection{Dataset Statistics for the Second Benchmark}
\label{subsec:dataset_stats}

In this section, we provide detailed statistics for the datasets used in our inductive link prediction benchmark.

\subsubsection{Statistics for the New Inductive Datasets}
Table~\ref{tab:new_inductive_datasets} reports statistics for CoDEx-M, WN18RR (Shomer et al. split), and HetioNet. 
For each dataset, we provide the number of edges, entities, and relations for both training and inference splits.

\begin{table}[h]
	\centering
	\caption{Statistics for the New Inductive Datasets.}
	\label{tab:new_inductive_datasets}
	\begin{tabular}{l l c c c}
		\toprule
		\textbf{Dataset} & \textbf{Split} & \# \textbf{triples} & \# \textbf{Entities} & \# \textbf{Relations} \\
		\midrule
		CoDEx-M & Train & 76,960 & 8,362 & 47 \\
		& Inference 1 & 69,073 & 8,003 & 40 \\
		\midrule
		WN18RR & Train & 24,584 & 12,142 & 11 \\
		& Inference 1 & 18,258 & 8,660 & 10 \\
		& Inference 2 & 5,838 & 2,975 & 8 \\
		\midrule
		HetioNet & Train & 101,667 & 3,971 & 14 \\
		& Inference 1 & 49,590 & 2,279 & 11 \\
		& Inference 2 & 37,927 & 2,455 & 12 \\
		\bottomrule
	\end{tabular}
\end{table}

\subsubsection{Statistics for Standard Inductive Benchmarks}
Due to space limitations in the main paper, we use abbreviated dataset names (e.g., W1 for WN18RR version 1). 
Table~\ref{tab:inductive_benchmark_stats} provides detailed statistics for multiple versions of WN18RR (W), FB15k-237 (F), and NELL-995 (N), including both source and target graphs.

\begin{table}[h]
	\centering
	\caption{Statistics of various versions of WN18RR (W), FB15k-237 (F), and NELL-995 (N).}
	\label{tab:inductive_benchmark_stats}
	\begin{tabular}{l c c c c c c}
		\toprule
		& \multicolumn{3}{c}{\textbf{Source KG}} & \multicolumn{3}{c}{\textbf{Target KG}} \\
		\cmidrule(lr){2-4} \cmidrule(lr){5-7}
		\textbf{Dataset} & \#\textbf{rel} & \#\textbf{ent} & \#\textbf{tri} & \#\textbf{rel} & \#\textbf{ent} & \#\textbf{tri} \\
		\midrule
		W1 & 9 & 2,746 & 6,678 & 8 & 922 & 1,618 \\
		W2 & 10 & 6,954 & 18,968 & 10 & 2,757 & 4,011 \\
		W3 & 11 & 12,078 & 32,150 & 11 & 5,084 & 6,327 \\
		W4 & 9 & 3,861 & 9,842 & 9 & 7,084 & 12,334 \\
		\midrule
		F1 & 180 & 1,594 & 5,226 & 142 & 1,093 & 1,993 \\
		F2 & 200 & 2,608 & 12,085 & 172 & 1,660 & 4,145 \\
		F3 & 215 & 3,668 & 22,394 & 183 & 2,501 & 7,406 \\
		F4 & 219 & 4,707 & 33,916 & 200 & 3,051 & 11,714 \\
		\midrule
		N1 & 14 & 3,103 & 5,540 & 14 & 225 & 833 \\
		N2 & 88 & 2,564 & 10,109 & 79 & 2,086 & 4,586 \\
		N3 & 142 & 4,647 & 20,117 & 122 & 3,566 & 8,048 \\
		N4 & 76 & 2,092 & 9,289 & 61 & 2,795 & 7,073 \\
		\bottomrule
	\end{tabular}
\end{table}

These datasets span a wide range of inductive scenarios, from minor shifts in entity sets to substantial structural differences between source and target graphs, enabling a comprehensive evaluation of inductive generalization.

\subsection{Computational Complexity Discussion}

The proposed framework employs a standard GCN on the model graph and an R-GCN on sampled subgraphs. Although both architectures follow a message-passing paradigm, their computational costs differ in structure.

For a standard GCN applied to a graph with $|\mathcal{V}|$ nodes and $|\mathcal{E}|$ edges, the per-layer complexity is:
\begin{equation}
	\mathcal{O}(|\mathcal{E}| d + |\mathcal{V}| d^2),
\end{equation}
where $d$ is the embedding dimension.

In contrast, R-GCN introduces relation-specific transformations. For a graph with $|\mathcal{R}|$ relation types, the per-layer complexity becomes:
\begin{equation}
	\mathcal{O}(|\mathcal{E}| d + |\mathcal{R}| |\mathcal{V}| d^2),
\end{equation}
due to the use of distinct weight matrices for each relation.

However, in our framework, R-GCN is applied only to small sampled subgraphs rather than the full knowledge graph. Since $|\mathcal{V}_s|$ and $|\mathcal{E}_s|$ are significantly smaller than those of the original graph, the overall computational cost remains manageable in practice.

Regarding the size of the model graph, each node represents a group of entities sharing identical relational feature vectors. Therefore, the number of model nodes satisfies:
\begin{equation}
	|\mathcal{V}_m| \leq |\mathcal{V}|.
\end{equation}

In practice, the model graph is often more compact than the original graph, as multiple entities may share the same relational patterns and are thus mapped to the same node. However, the degree of this compression depends on the diversity of relational features in the dataset. In cases where entity features are highly unique, the size of the model graph may approach that of the original graph.


\end{document}